\documentclass[12pt]{article}

\usepackage{scicite}

\usepackage{times}
\usepackage{graphicx} 

\usepackage{nth}
\usepackage{xcolor}
\usepackage{amsmath} 
\usepackage{booktabs}
\usepackage{longtable}

\usepackage{multirow}  

\usepackage{chngcntr} 

\usepackage{hyperref}

\newenvironment{sciabstract}{%
\begin{quote} \bf}
{\end{quote}}

\newcounter{lastnote}

\title{AI for operational methane emitter monitoring \\from space}

\author{Anna Vaughan\textsuperscript{1*†}, Gonzalo Mateo-Garcia\textsuperscript{1*†}, Itziar Irakulis-Loitxate\textsuperscript{1,4†},  \\ Marc Watine\textsuperscript{1}, Pablo Fernandez-Poblaciones\textsuperscript{1}, Richard E. Turner\textsuperscript{2}, \\
James Requeima\textsuperscript{3}, Javier Gorroño\textsuperscript{4}, Cynthia Randles\textsuperscript{1}, \\ Manfredi Caltagirone\textsuperscript{1} and Claudio Cifarelli\textsuperscript{1*}\\
\\
\small{\textsuperscript{1}International Methane Emissions Observatory, United Nations Environment Programme}\\
\small{\textsuperscript{2}Department of Engineering, University of Cambridge}\\
\small{\textsuperscript{3}Vector Institute, University of Toronto}\\
\small{\textsuperscript{4}Research Institute of Water and Environmental Engineering (IIAMA)},\\ \small{Universitat Politècnica de València (UPV)}\\
\small{\textsuperscript{*}To whom correspondence should be addressed; E-mail:  anna.vaughan@un.org},\\ \small{ gonzalo.mateogarcia@un.org and claudio.cifarelli@un.org}\\
\small{\textsuperscript{†} These authors contributed equally to this work}}

\date{25/7/2024}

\begin{document} 




\maketitle


\begin{sciabstract}
Mitigating methane emissions is the fastest way to stop global warming in the short-term and buy humanity time to decarbonise. Despite the demonstrated ability of remote sensing instruments to detect methane plumes, no system has been available to routinely monitor and act on these events. We present MARS-S2L, an automated AI-driven methane emitter monitoring system for Sentinel-2 and Landsat satellite imagery deployed operationally at the United Nations Environment Programme’s International Methane Emissions Observatory. We compile a global dataset of thousands of emission events for training and evaluation,  demonstrating that MARS-S2L can skillfully monitor emissions in a diverse range of regions globally, providing a 216\% improvement in mean average precision over a current state-of-the-art detection method. Running this system operationally for six months has yielded 457 near-real-time detections in 22 different countries of which 62 have already been used to provide formal  notifications to governments and stakeholders. 
\end{sciabstract}

\section*{Introduction}

Hundreds of miles offshore in the Gulf of Thailand, a malfunctioning flare on an oil platform is emitting vast amounts of methane into the atmosphere. A potent greenhouse gas, methane is responsible for more than 25\% of global warming to date ~\cite{stocker2014climate,varon2021high,dean2018methane} and mitigating emissions is the fastest way to reduce the worst effects of climate change in the short-term and buy humanity time to decarbonize. Discovered in 2023 \cite{valverde2024satellite}, a time-series analysis of satellite images recently revealed that this platform has been frequently emitting methane at a rate of several tonnes every hour for over a decade without detection or intervention \cite{clark2024big} (see Figure~\ref{fig:malaysia} materials and methods). \\

This site and many others like it around the world are examples of methane emitters.
Unlike other greenhouse gasses such as carbon dioxide, a relatively small number of these point sources make up a significant percentage of total emissions, presenting a crucial opportunity for rapid mitigation~\cite{lauvaux2022global}. Remarkably, despite their climate impact and increasing public awareness after events such as the Nordstream pipeline leak \cite{jia2022nord,unep2023nordstream} or the record-breaking Kazakhstan blowout~\cite{guanter_multisatellite_2024}, systematic documentation of and regular monitoring of these sites has not been available due to challenges detecting methane in satellite images in an automated manner. \\

Although a wide variety of satellites are suitable for detecting methane, currently the only platforms with sufficiently high revisit time and spatial resolution for regular monitoring and attribution of emissions are Sentinel-2 and Landsat. Together, these satellites provide publicly available imagery of all land surface every 2.3 days on average at 20-30 meters spatial resolution and are capable of detecting emissions at a rate of as little as 1000 kg/hour \cite{ehret_global_2022,gorrono2023understanding} in favourable conditions. Unfortunately, the spectral resolution of these instruments is low, making detecting the methane signal a challenging task. Early work in this area utilised simple image thresholding techniques \cite{varon2021high,irakulis2022satellites,gorrono2023understanding}, requiring time-consuming manual checking and achieving limited accuracy. In recent years, there has been significant interest in using machine learning to detect and monitor methane emissions \cite{ruuvzivcka2023semantic,vaughan2024ch4net,schuit2023automated,joyce_using_2023,radman_s2metnet_2023,bruno_u-plume_2024,mehrdad2021prediction,jongaramrungruang2022methanet,rouet2024automatic}. For multispectral imagery, CH4Net \cite{vaughan2023ch4net} provided the first AI model for fully automated monitoring of methane emission events at a set of locations in Turkmenistan using Sentinel-2 images, providing orders of magnitude improvement over previous techniques when evaluated on several hundred real emission events.\\

These promising early results have led several authors to suggest that these methods could be used to build a global, fully automated operational monitoring system for methane emissions \cite{vaughan2023ch4net,rouet2024automatic,ruuvzivcka2023semantic}. Despite the urgency involved in addressing these issues, implementation of such a system has not yet been realised due to the significant challenges involved. The lack of a large global dataset of real, verified methane plumes in multispectral data, machine learning challenges in detecting weak signals in multispectral images, developing operational infrastructure for analysis of results, and facilitating engagement with governments and asset owners are all major roadblocks to developing such a system. Here, we report that this objective has now been accomplished. We present MARS-S2L, an automated AI driven methane emitter monitoring system developed as part of the Methane Alert and Response System (MARS) at the United Nations Environment Programme's International Methane Emissions Observatory. Launched at COP27, MARS is the first global satellite detection and notification system for methane emissions, for which MARS-S2L is a key element. By compiling a large hand-annotated global dataset of methane emission events, training a novel AI model, and building a quality-assurance and quality control (QA/QC) tool to easily interpret and act on alerts, we create a system capable of automatically monitoring methane emitters globally. We report results for the first six months of operation, demonstrating skillful detection of 457 emissions and the creation of 62 formal notifications.

\section*{MARS-S2L}

\subsection*{\textit{Dataset}}

The first step in using AI for methane emitter detection is building a large dataset of events for training and evaluation, which we name the MARS-S2L dataset. As methane emission events are localised and relatively rare, this presents a significant challenge, with many previous studies forced to examine only limited areas \cite{vaughan2023ch4net}, and rely on synthetic data \cite{joyce_using_2023,radman_s2metnet_2023,rouet2024automatic}, or data taken days before a ground truth plume detection \cite{rouet2024automatic}. To ensure the trustworthiness and suitability for operational usage of a global monitoring system, it is necessary to have a large dataset of real plumes in multispectral imagery verified by experts. The MARS-S2L dataset utilises observations from four sensors: the Multispectral Imager (MSI) aboard Sentinel-2 A and B \cite{drusch2012sentinel} and the Operational Land Imager (OLI) aboard Landsat 8 and 9 \cite{wulder2019current}. Together these platforms have a mean revisit time of 2.3 days \cite{li2020global}. \\

By manually inspecting known emission sites identified by IMEO analysts from scarce but highly accurate hyperspectral imagery, we compile a large dataset of 53,309 images, including 4,230 emissions over 707 distinct emitter sites in a diverse range of regions globally between January 2018 and June 2024 (Figure 1 (b)). For each image, a multi-band multi-pass image~\cite{irakulis2022satellites} is created using the most similar previous overpass in the non-methane bands and manually checked for methane plumes. If a emission is present a mask is hand annotated on the image, a challenging task as the plume is frequently weak and poorly defined (Figure 1 (a)). We note that the resulting dataset is heterogeneous as not every image is labelled for every site. For a detailed exposition of the dataset see materials and methods.\\

\begin{figure}
     \centering
     \includegraphics[width=1.02\linewidth]{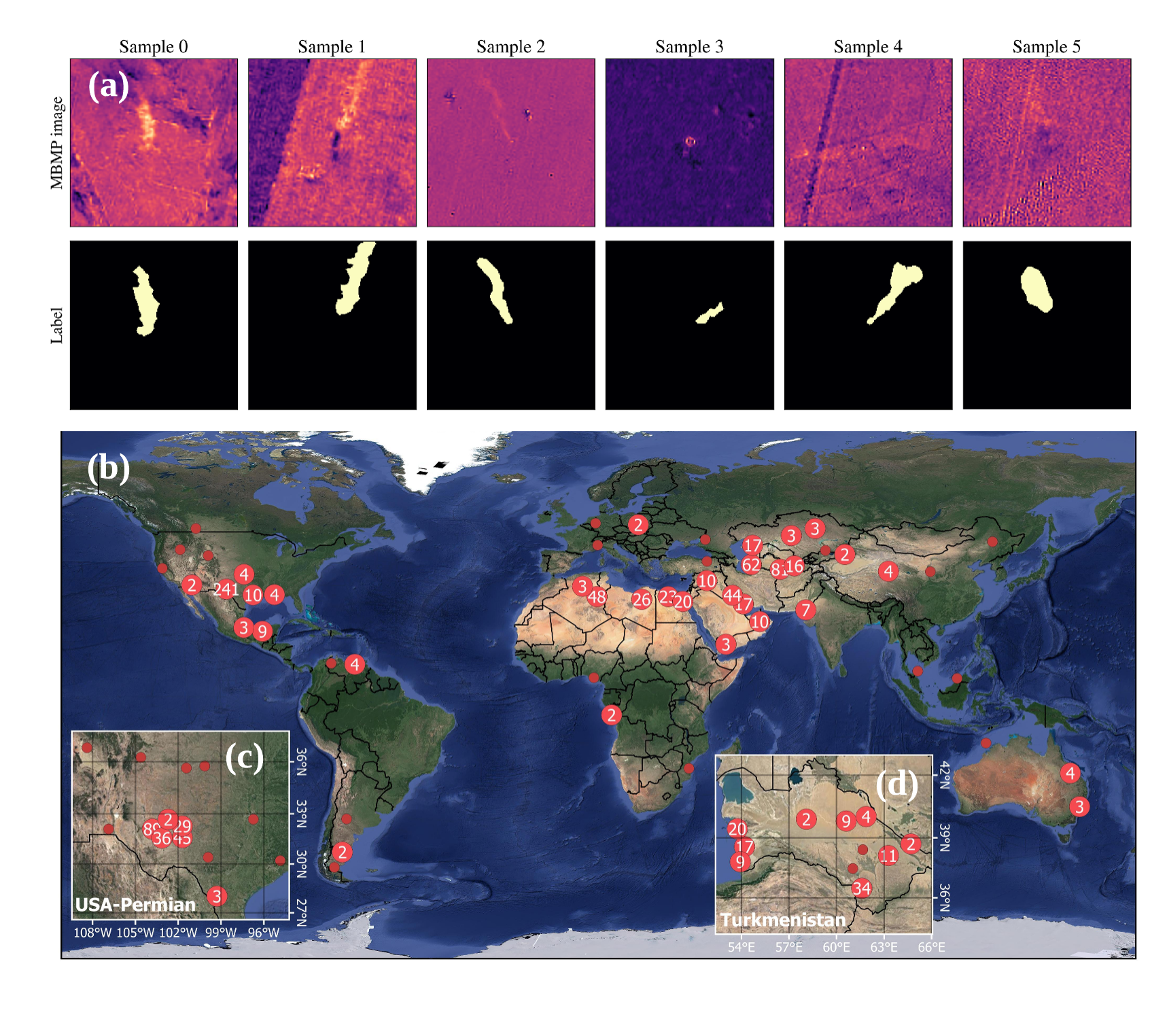}
     \caption{\textit{\textbf{Characteristics of the MARS-S2L dataset} In total 53,309 images containing 4,230 emissions over 707 distinct emitters globally are included in the dataset. Maps show the locations of (b) all included sites globally, the Permian Basin (c) and Turkmenistan (d). Examples of plumes in multiband-multipass imagery together with the corresponding hand-annotated masks are shown in (a) with three examples of clear plumes (Samples 1,2,3) and three examples of poorly-defined plumes (samples 4,5,6).}}
     \label{fig:1}
\end{figure}

\subsection*{\textit{AI model}}

 MARS-S2L is implemented as a convolutional conditional neural process \cite{gordon2019convolutional} with film layers for parameter-efficient finetuning \cite{requeima2019fast,perez2018film}. Neural processes are a class of models ideally suited to probabilistic data fusion tasks such as methane detection as they are capable of handling missing data and providing probabilistic output, and have been applied extensively to environmental applications \cite{vaughan2022convolutional,andersson2022active,bruinsma2023autoregressive,markou2022practical,vaughan2024aardvark}. As multispectral methane retrieval performance is strongly dependant on background surface characteristics \cite{gorrono2023understanding}, the addition of a set of film parameters for each location allows the model to learn site-specific features and for rapid adaptation to new sites added to the system without needing to retrain the entire model. Inputs to MARS-S2L are six multispectral bands common to Sentinel-2 and Landsat for both the current pass and the most similar cloud free pass from the past 3 months. Auxiliary data includes cloud masks generated using CloudSEN12 \cite{aybar_cloudsen12_2022} and northward and eastward 10m wind from ERA5-Land reanalysis \cite{munoz2021era5} or GEOS-FP for offshore platforms \cite{lucchesi2013file}. We also include multi-band multi-pass (MBMP)~\cite{irakulis2022satellites} differencing of the two multispectral passes as an auxiliary channel. Each prediction consists of a probabilistic plume mask, which is converted to a scene-level probability by thresholding the per-pixel values (see materials and methods). A schematic of the model architecture is shown in Figure 2 (a). MARS-S2L is trained  on all images prior to December 2023, with 2021 held out as a validation year comprising a total of 34,081 training images over 532 sites including 3,123 plumes. For sites with a sufficiently large number of training images a set of FiLM parameters is learned, with an extra set learned to use for all sites where insufficient data is available. The model is evaluated on all data from December 2023 to June 2024, comprising a total of 13,448 test images over 680 sites including 829 plumes. For details of training see materials and methods. 


\subsection*{\textit{The PlumeViewer}}

For MARS-S2L to be useful in an operational setting, results need to be displayed in a manner that analysts can easily obtain, explore, and interpret new predictions and use these to engage directly with governments and operators. To facilitate this process we developed the PlumeViewer, a QA/QC tool that displays model outputs and past predictions over the same location. Each morning at 6:30 GMT, any new Sentinel-2 and Landsat image covering a location in monitoring is ingested in the system and processed by the MARS-S2L model. When analysts log into the PlumeViewer, the alert screen with the latest positive model predictions are shown with filters in the probability score, fluxrate, satellite or country (Figure 2 (b)). The analyst can then click on any of these alerts to inspect and validate the predicted plume, get the emission flux quantification and error estimation, and cross-reference the prediction with different infrastructure and proprietary databases, as well as possible previous detections from these or other satellites to which IMEO has access. As many images as possible are processed each day depending on analyst availability and the time taken on each case, hence there is no hard probability cutoff for an image to be inspected. If the source of the emission and asset owner is identified, the information collected from the satellites will be shared with the government and the company contact, through the MARS regional case manager. For detailed images of the analyst process in the PlumeViewer see materials and methods. \\

\begin{figure}
     \centering
     \includegraphics[width=1.02\linewidth]{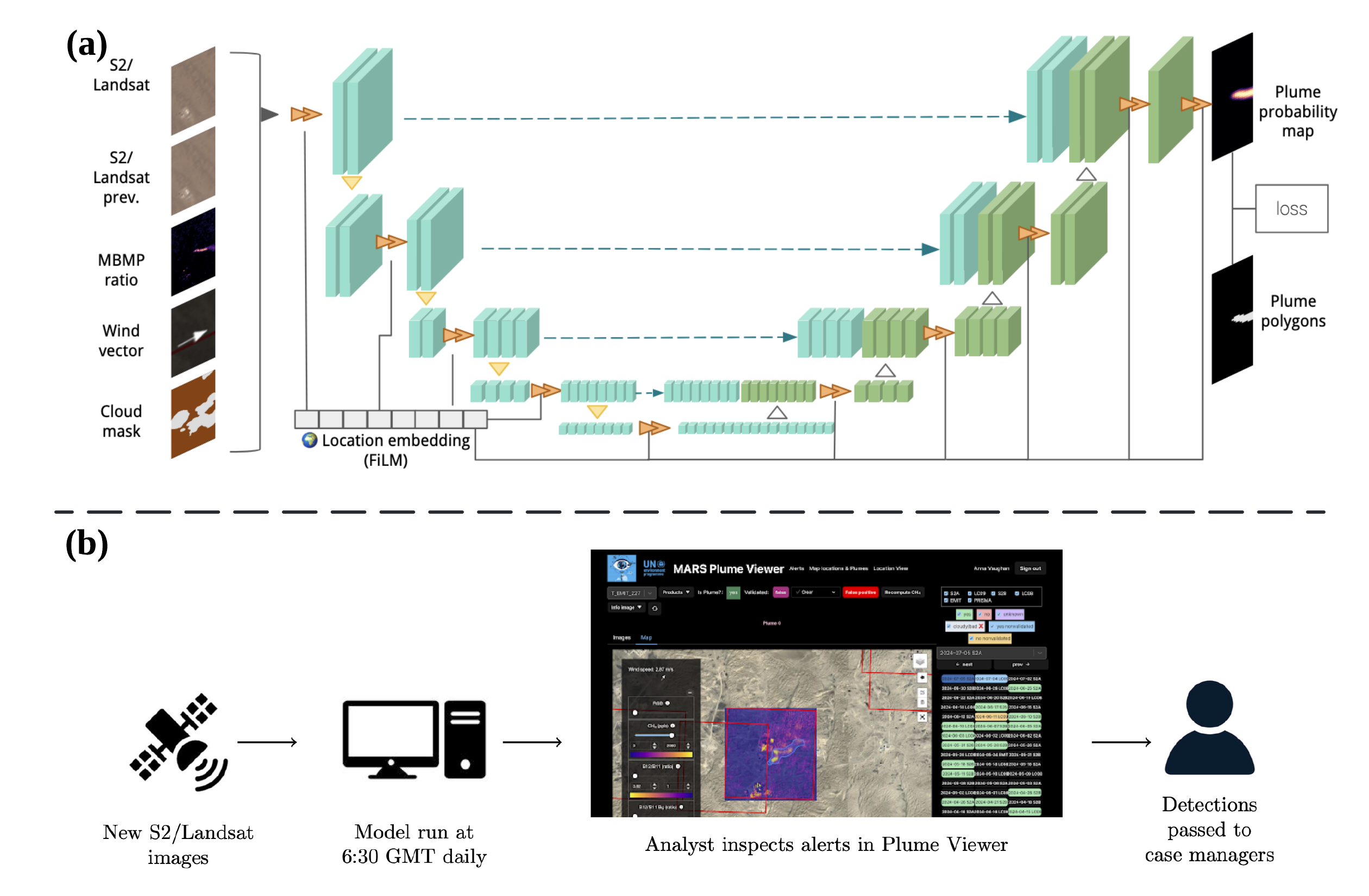}
     \caption{\textit{\textbf{MARS-S2L model architecture and deployment} (a) the architecture and inputs of the MARS-S2L model. The multispectral bands from the current and previous overpass the site together with the MBMP image and wind information are used as inputs. A cloud mask is first generated using the CloudSEN12 model. All data is then fed into MARS-S2L which outputs the probability that each pixel is part of a methane plume. (b) shows the operational deployment process. At 06:30 every morning any new Sentinel-2 and Landsat images are downloaded and predictions generated. These are then shown in the PlumeViewer where analysts inspect each alert and provide details of events over known assets to case managers to issue notifications.}}
     \label{fig:2}
\end{figure}

\section*{Results}

We conduct two analyses of results. The first is to present the results of running MARS-S2L operationally in near real-time over a period of six months. As analysts do not manually verify every image available, checking only those with high probabilities assigned, it is only possible to report the number of true positives not true negatives and false positive rates. We therefore conduct a rigorous systematic evaluation of performance over all labelled data in the next section. \\

\subsection*{Near real time deployment results}

In its current operational configuration MARS-S2L provides daily monitoring of a set of 707 longitude-latitude locations where methane emissions have been verified to occur by IMEO analysts. This list of locations is frequently updated as further sites are identified. We report results from running MARS-S2L for a six-month trial period from the 1st of January 2024 to the 30th of June 2024. During this phase, the model produced 457 detections from 110 unique sites globally in 22 different countries. Already during this time 62 emissions detected at 41 unique sites in 15 different countries were notified to government and corporate stakeholders. Examples of three notified plumes are shown in Figure 3 (c-e). As notifications can only be issued for locations where IMEO has established a point of contact, the percentage of notified detections is expected to rapidly increase as engagement with governments and operators deepens in the second half of 2024. \\

\begin{figure}
     \centering
     \includegraphics[width=1.02\linewidth]{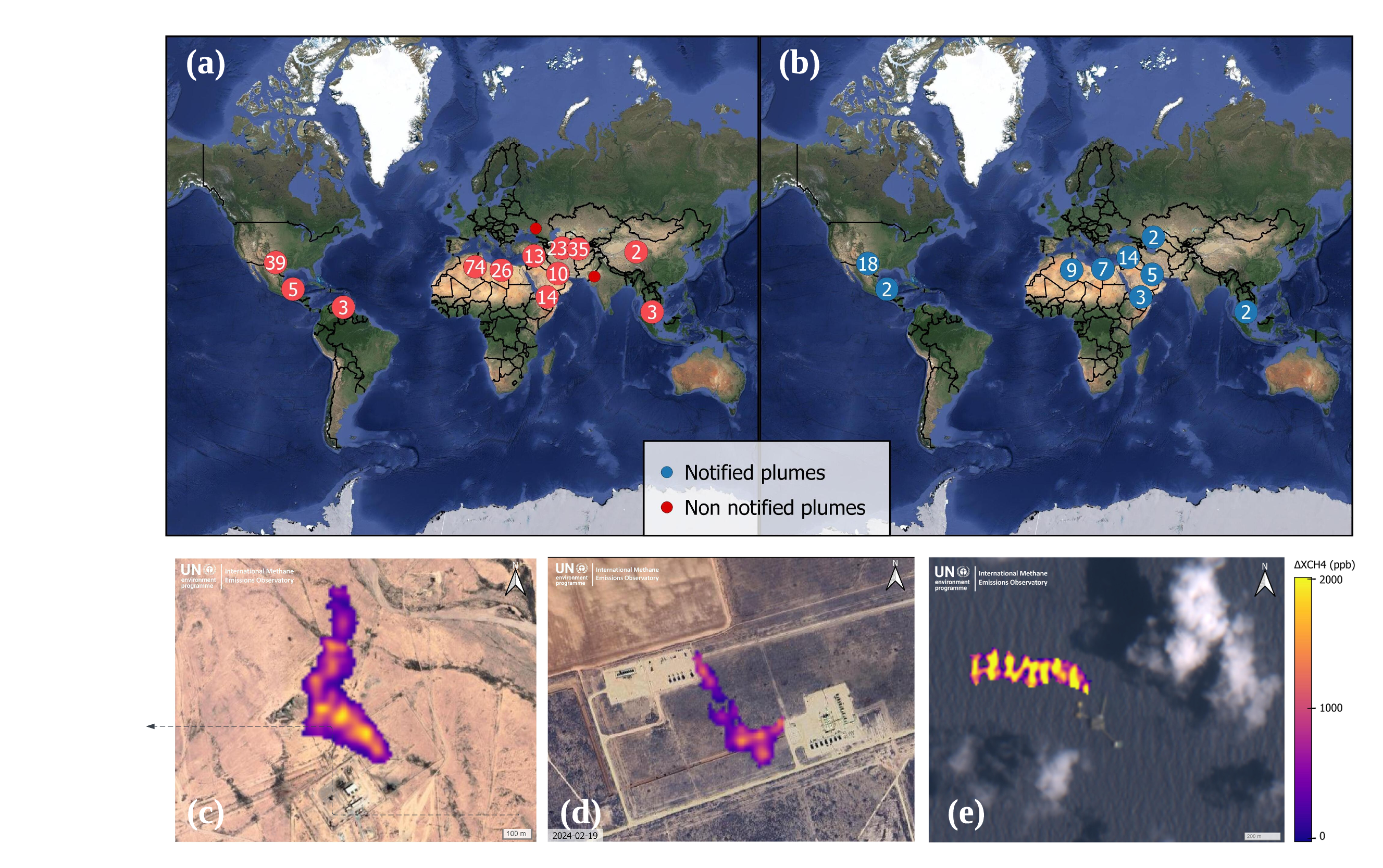}
     \caption{\textit{\textbf{Model detections during six months of operational deployment.} MARS-S2L detections are shown for (a) non-notified plumes and (b) notified plumes. Examples of three notified events are shown in for plumes in (c) the US, (d) Syria and (e) Thailand. MARS-S2L successfully identified 457 emissions in near real time leading to 62 notifications.}}
     \label{fig:2}
\end{figure}

\subsection*{Global evaluation of performance}
We conduct a detailed evaluation of the performance of MARS-S2L on the held out test set. These sites primarily consist of oil and gas production facilities, together with several coal facilities and landfill sites. We compare two models: the previous state-of-the art AI model for methane monitoring, CH4Net~\cite{vaughan2024ch4net}, and MARS-S2L. Figure 4 (a) compares the performance of CH4Net and MARS-S2L for plumes stratified by flux rate. MARS-S2L substantially outperforms CH4Net, especially for the large emissions of most interest for mitigation work. MARS-S2L is also able to successfully identify several very small emissions in the range of 0.5-1 tonnes/hour. An example of a MARS-S2L prediction for an event in Algeria is shown in Figure 4 (b-e). For further examples of model predictions see materials and methods. MARS-S2L achieves a mean average precision of 0.67 compared with 0.31 for CH4Net, a 216\% improvement over the previous state-of-the-art. Taking a threshold of 0.5 for binary classification, MARS-S2L achieves an accuracy of 0.90, recall of 0.77, precision of 0.37, and 9\% false positive rate, compared to 0.91, 0.42, 0.32 and 6\% for CH4Net.  We note that these binary classification results are provided only to give an overview of model performance on these metrics, in practice, there is no hard probability cutoff for operational inspection with analysts processing as many images as they have time for each day. Evaluation on the Stanford controlled release experiments~\cite{sherwin_single-blind_2023,sherwin_single-blind_2024} demonstrates similar performance and the ability to detect small emissions (see materials and methods). \\

\begin{figure}
     \centering
     \includegraphics[width=1.02\linewidth]{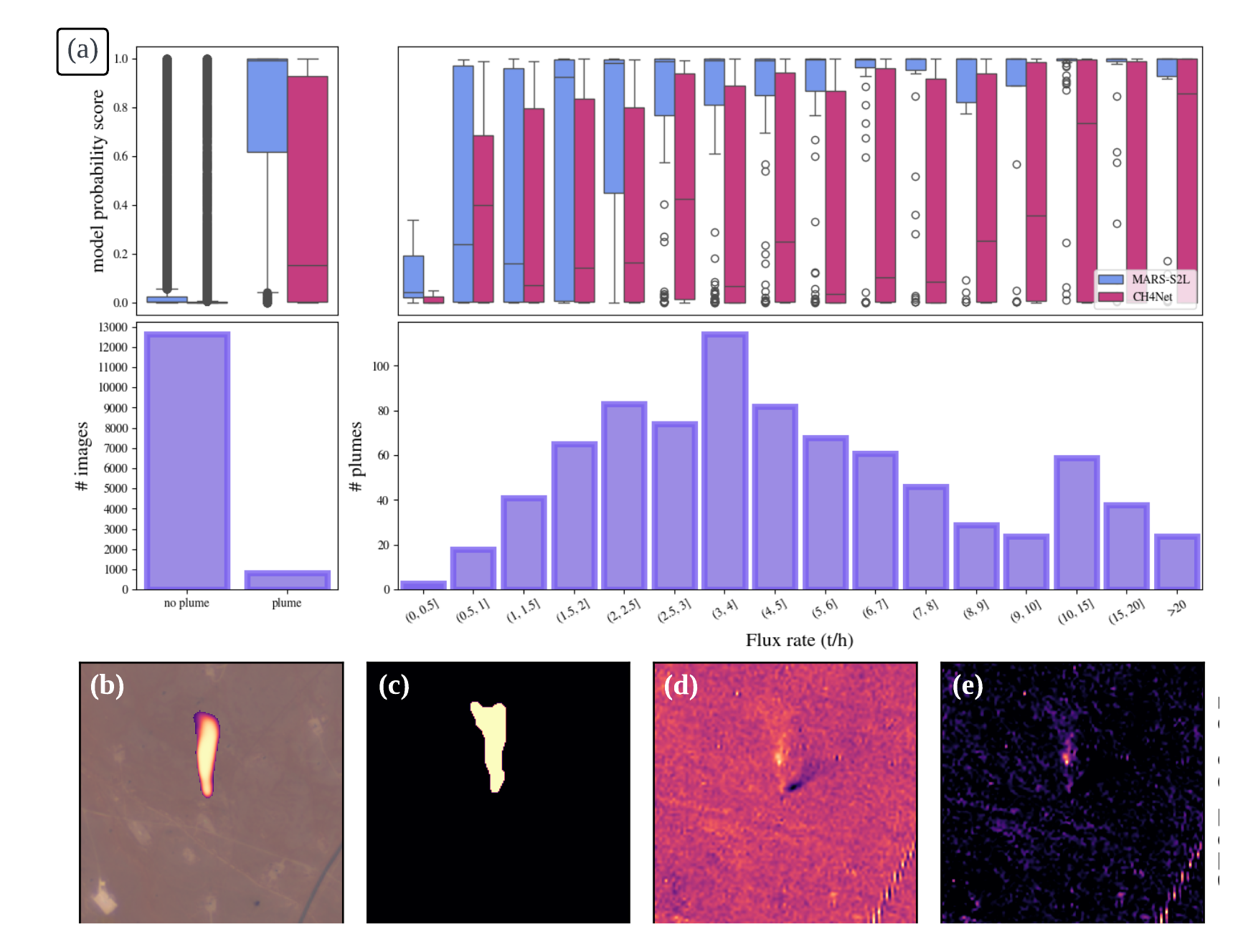}
     \caption{\textit{\textbf{Global performance results} Global results showing (a) Performance of MARS-S2L and CH4Net as a function of flux rate. MARS-S2L achieves excellent performance and substantially outperforms CH4Net. An example of successful plume identification for a site in Algeria is shown below, with (b) predicted probability superimposed on RGB imagery, (c) corresponding hand-annotated mask, (d) multiband-multipass image and (e) CH4 enhancement. }}
     \label{fig:3}
\end{figure}

\subsection*{Case studies}
We next evaluate performance in different regions by conducting three case studies of areas with different background characteristics. These are Turkmenistan, a desert area ideally suited to methane detection with multispectral imagery, the Permian Basin, an oil and gas production region in the South Western United States with more challenging background surfaces and offshore platforms. This explores the ability of MARS-S2L to identify plumes in regions with a diverse range of background characteristics. \\

MARS-S2L achieves excellent performance over Turkmenistan with accuracy over all images of 0.88, mean average precision of 0.74 and false positive rate of 10\% (Figure 5 (b)). Spatial predictions for a time-series of emission events provide accurate predictions of plume morphology and extent across a range of different events. The Permian Basin is a substantially more challenging region, with MARS-S2L providing predictions with a mean average precision of 0.54, accuracy of 0.95 and false positive rate of 5\% (Figure 5 (a)). For the offshore platforms, the model achieves a mean average precision of 0.68, accuracy of 0.76 and false positive rate of 26\%, indicating skillful performance in a region with very different background characteristics from the majority of the training data (Figure 5 (c)). We note that these binary classification results could likely be improved by setting a separate probability threshold on a per-site or per-region basis. Taken together, these three case studies indicate that MARS-S2L is capable of skillfully detecting emissions in a variety of regions globally with different background types.

\begin{figure}
     \centering
     \includegraphics[width=1.02\linewidth]{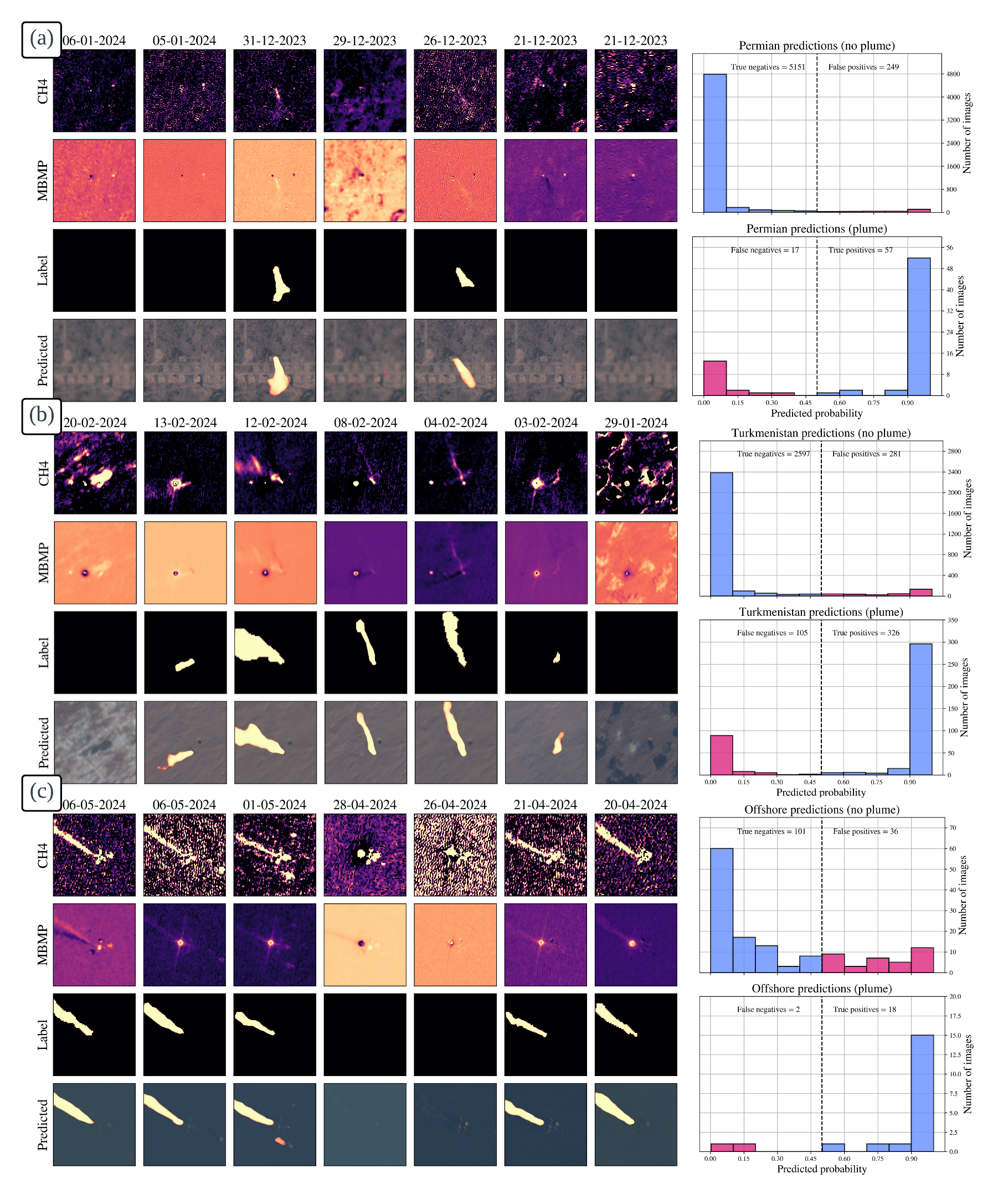}
     \caption{\textit{\textbf{Case study results.} Case study results for (a) the Permian Basin, (b) Turkmenistan and (c) offshore platforms. For each case an example time-series of predictions for one site is shown on the left with  CH4 enhancement, multi-band multi-pass image, hand-annotated mask and model prediction. Histograms of model predictions over all sites for images with and without a plume are shown on the right. For all three cases MARS-S2L provides skillful identification of plumes.}}
     \label{fig:4}
\end{figure}

\section*{Discussion}

We have presented MARS-S2L, the first AI-driven operational methane emitter monitoring system. By compiling a large, expert-verified global dataset of methane emission events we train a novel machine learning model that outperforms a previous state-of-the-art system by large margins and automatically detects plumes globally in very diverse regions. Running this model operationally for six months, hundreds of emission events were successfully detected in near-real-time in 22 different countries of which 62 have been utilised to send notifications to parties with the power to act on and mitigate these emissions. \\

MARS-S2L is having significant and immediate impacts on the methane emitter mitigation work, allowing a small group of analysts at IMEO to operationally exploit the wealth of data from Sentinel-2 and Landsat-8/9 to notify and engage with relevant stakeholders when methane leaks are detected, and drive diplomatic and policy efforts to stop these damaging events. All detected plumes (notified or not) from MARS-S2L are published in a public portal at methanedata.unep.org on a bi-weekly basis. These detections are used by a broad cross-section of society, including the energy industry, policymakers, scientists from directly and indirectly related research fields, NGOs, media communicators, private companies, and others. In addition, IMEO will be releasing the MARS-S2L model and dataset described in this paper in machine-learning ready format creating a consistent, verified database for model training, evaluation and inter-comparison to spur cohesive research efforts in this critical field.\\

There is a rapidly approaching satellite data revolution for methane detection, marked by the launch of several highly anticipated instruments such as Carbon Mapper and MethaneSat \cite{rohrschneider2021methanesat}. To fully realize the potential of these instruments for methane mitigation, a pipeline as described in this paper will be instrumental as even an army of analysts could not fully exploit the expected data volume. MARS-S2L is easily extended to new sensors, and we anticipate that this will be the first of multiple initiatives using AI to monitor and detect methane emissions and greenhouse gas emissions more broadly in operational settings. Orthogonal to this, there is substantial potential to iterate on the machine learning architecture proposed here, with rapid advances in AI for remote sensing occurring even during the period of this study \cite{hong2024spectralgpt,cha2023billion}. A second generation model utilising recent machine learning advances to both monitor and detect emissions from multiple sensors will be deployed at IMEO in the second half of 2024, further pushing the boundaries of AI in this crucial field. \\

\section*{Data and code availability}
Model weights, code, and data will be made publicly available under a Creative Commons license CC BY-NC-SA 4.0 DEED (Attribution-NonCommercial-ShareAlike 4.0 International) on completion of the peer review process. This will be provided in a machine learning ready format to drive coherent  research in this field. 

\section*{Acknowledgements}
The authors thank Meghan Demeter, Giulia Bonazzi, Tharwat Mokalled, and Florencia Carreras as the MARS members in charge of country engagement and dealing with the notification process, as well as the rest of the IMEO group, including the OGMP 2.0 group, and the EDF for their support, advice and fruitful discussions. We thank Luis Gomez-Chova, Cesar Aybar and Enrique Portales-Julia for adapting the CloudSEN12 cloud detection model to Landsat and for the support of georeader package. We also thank the members of the Science Oversight Committee for their feedback. We thank Luis Guanter and the rest of the team at the Universitat Politècnica de València (UPV) for their support of this work.

\section*{Author contributions}
A.V, C.C and I.I-L conceptualised the project. A.V collected and labelled the training dataset, designed the AI model and developed the initial version. G.M-G developed the PlumeViewer and operational deployment pipeline with assistance from P.F-P. A.V and G.M-G developed the final version of the AI model. I.I-L lead the identification of new sites and development of the evaluation dataset and notification process, assisted by M.W, G.M-G and A.V. C.C coordinated the project. R.E.T and J.R provided assistance with machine learning techniques. J.G provided assistance with the physics-based simulation scheme. A.V and G.M-G wrote the initial draft of the manuscript. C.R, M.C and all aforementioned authors provided feedback on results throughout the project and contributed to the final version of the manuscript.

\bibliography{scibib}

\bibliographystyle{Science}

\clearpage

\section*{Materials and methods}

\counterwithin*{figure}{section}
\renewcommand{\thefigure}{S\arabic{figure}}
\renewcommand{\thetable}{S\arabic{table}}

\setcounter{figure}{1} 
\setcounter{table}{1} 

\subsection*{Input datasets}

Sentinel-2A and Sentinel-2B are two Earth observation satellites that are part of the Copernicus program, coordinated by the European Commission in partnership with the European Space Agency (ESA). Launched in June 2015 (Sentinel-2A) and March 2017 (Sentinel-2B), these platforms capture high-resolution images in 13 spectral bands using the Multispectral Imager (MSI) instrument. Landsat 8 and Landsat 9 are part of the Landsat program, a joint initiative by NASA and the United States Geological Survey (USGS) that has been observing Earth's landmasses since 1972. Launched in February 2013 and September 2021, respectively, these satellites are equipped with advanced sensors, including the Operational Land Imager (OLI) and the Thermal Infrared Sensor (TIRS), which collect imagery in multiple spectral bands. \\

MARS-S2L is trained using the following common bands of OLI and MSI: blue (MSI: ~490 nm, OLI: ~482 nm), green (MSI: ~560 nm, OLI: ~561 nm), red (MSI: ~665 nm, OLI: ~655 nm),  NIR (MSI: ~842 nm, OLI: ~865 nm) and both SWIR bands at wavelengths ~1610 nm and ~2190 nm. We exclude aerosol and cirrus bands of both sensors as they have significantly different spatial resolution.  Bands of both sensors with spatial resolution higher than 10m are interpolated to 10m using bicubic interpolation. Although there are slight differences in central wavelength, band width and resolution between both sensors, these differences are not sufficient to impact model performance. \\

Several auxiliary data channels are provided. For each image cloud and cloud shadow masks were computed using the CloudSEN12 model \cite{aybar_cloudsen12_2022}. Wind vectors are obtained from the ERA5-Land reanalysis \cite{munoz2021era5} and NASA GEOS-FP \cite{lucchesi2013file} for offshore locations.

\subsection*{Image processing, labelling and quantification}
To train and evaluate MARS-S2L we create a comprehensive database of methane plumes building on a list of known locations with past detected emissions provided by the International Methane Emission Observatory (IMEO) analysts. For each of these locations Sentinel-2 and Landsat images are downloaded for a 2$\times$2 km$^2$ square around the source. Each image is  processed with CloudSEN12 and discarded if it contains more than 50\% cloud, cloud shadow or missing pixels. For clear images, the retrieval is computed using the multi-band multi-pass (MBMP) ratio \cite{irakulis2022satellites} using the most similar cloud-free image. We measure similarity in the visible and SWIR1 band and restrict to images acquired over the last 4 months. For offshore platforms, we use the multi-band single-pass ratio (MBSP)~\cite{irakulis-loitxate_satellites_2022}. From the retrieval image, plumes are manually annotated for each image using the Computer Vision Annotation Tool ~\cite{CVAT_2023} and labelling functionality within PlumeViewer. For positive examples we estimate the per-pixel concentration of methane and the flux rate in kilograms per hour (kg/h) following Gorroño et al.~\cite{gorrono2023understanding}.    \\

\subsection*{Dataset}
In total, the MARS-S2L dataset compiled as part of this work comprises images over 707 emitters in 33 countries. A large percentage of total images and observed plumes are from known oil and gas producing hotspots in Algeria, Turkmenistan, Iraq, Kazakhstan and Libya, with smaller numbers in other locations. Table \ref{tab:stats} and Table \ref{tab:stats_test} show the numbers of locations, images and plumes per country for the complete dataset and test dataset respectively. The number of images, locations and plumes analysed over each country per month for 2023 and 2024 is shown in Figures \ref{fig:nimagesplumesbycountrymars2} and \ref{fig:nimagesplumesbycountryrest}. Total numbers of images and plumes stratified by training split and satellite are shown in Figure \ref{tab:splits}.

\subsection*{Model design and architecture}

The model is implemented as a convolutional conditional neural  process \cite{gordon2019convolutional} with FiLM layers for finetuning inspired by \cite{requeima2019fast}, with the density channel giving locations of invalid pixels provided by Cloud SEN12. MARS-S2L is trained to maximise the log likelihood of a Bernoulli distribution at each pixel with positive pixels upweighted by a factor of 10 to account for the pixel-wise class imbalance. The backbone decoder is implemented as a simple and flexible UNet architecture \cite{ronneberger2015u} consisting of four encoder blocks (2D convolution layer, batch norm, ReLU activation, max pool) followed by four decoder blocks (transposed 2D convolution layer, 2D convolution layer, batch norm, ReLU activation, 2D convolution layer, batch norm ReLU activation) with skip connections between blocks of corresponding scale. Channel output dimensions for each of these blocks are with kernel sizes of 3 for all convolution layers and 2 for the max pooling layers.\\

For each site with at least five positive images available we learn a set of parameters specific to each location. Feature-wise Linear Modulation layers (FiLM layers) are a lightweight method for finetuning convolutional architectures \cite{perez2018film}. FiLM layers dynamically adjust the activations of convolutional feature maps by modulating them with learned parameters. Mathematically, this is expressed as $\gamma \times x + \beta$ where $\gamma$ and $\beta$ are learned parameters of an affine transformation for each convolutional filter and $x$ is the input feature map. By applying theses transformations independently to each channel of the feature maps, FiLM layers allow the network to selectively amplify or attenuate specific features depending on scene-specific characteristics.  


\subsection*{Model training procedure}

\begin{figure}
    \centering
    \includegraphics[width=0.8\linewidth]{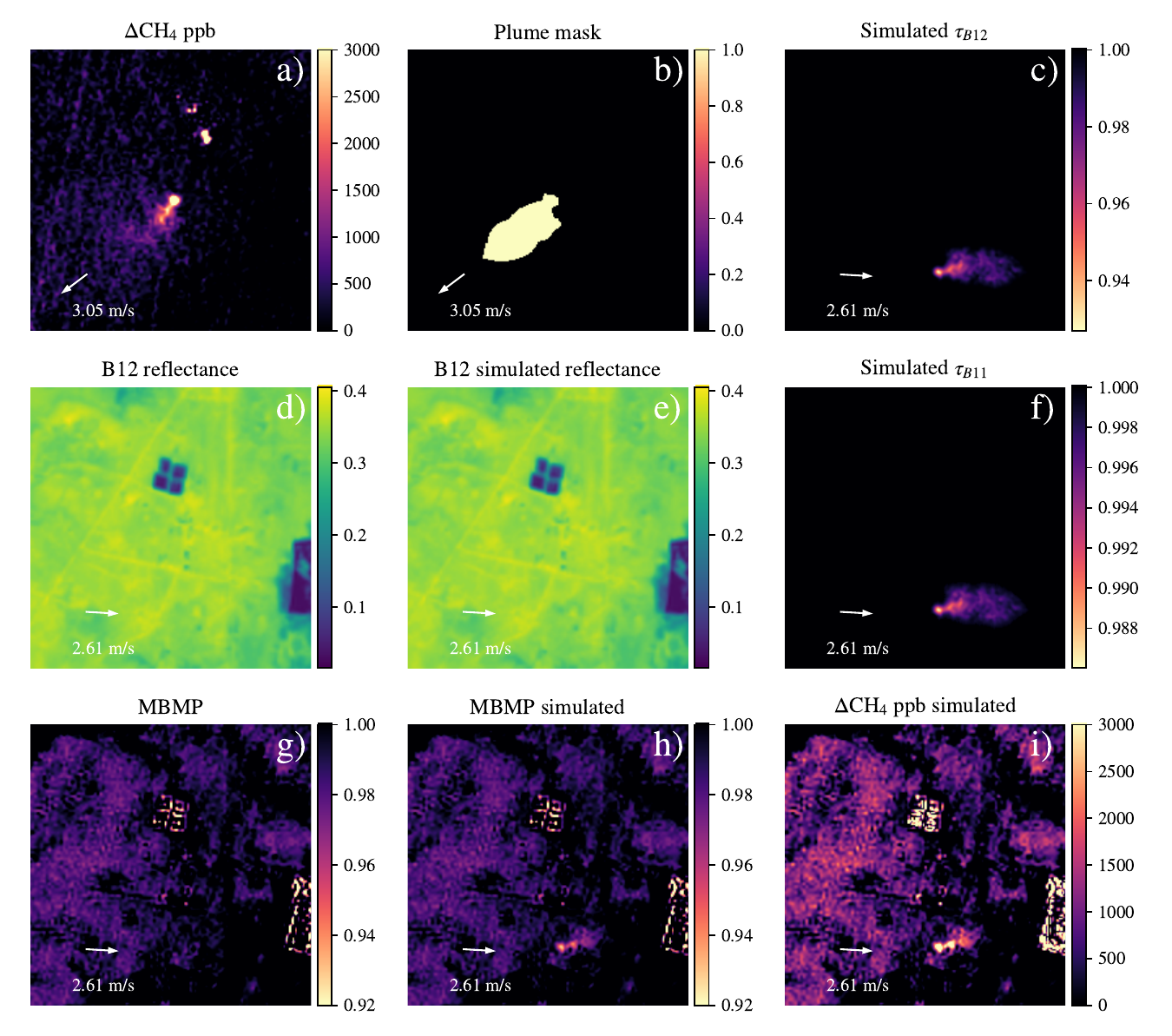}
    \caption{Samples highlighting the simulation procedure: (a) quantified retrieval $\Delta$CH$_4$ of a plume, (b) binary plume mask for the same plume, (d) B12 reflectance and (g) MBMP for an image in a different location without a plume. Transmittance of the plume in bands 11 (f) and 12 (c) aligned with the wind of the clear location, B12 band of the clear image with the simulated plume (e), MBMP retrieval (h) and quantified retrieval (i).}
    \label{fig:simulation_images}
\end{figure}

Training is performed using stratified sampling and a novel physics-based simulation scheme (Figure ~\ref{fig:simulation_images}) designed to compensate for the dataset imbalance of plumes and images across locations. For each call to the dataloader we randomly sample a location, and a binary indicator denoting whether we are sampling an image with or without a plume. As it is desirable to train on real images where possible, plumes are simulated depending on the number of real plumes available at each location. For locations with no real plumes, we randomly sample a negative image and simulate a plume following the procedure below. In the case of one to five real plumes, a synthetic image is created with probability 0.9 and a real image is used otherwise. Finally, in cases where a site has more than five images with real plumes available, a synthetic image is created with probability 0.1 and a real image used otherwise. Although these thresholds are arbitrarily selected, changing these probability thresholds was found to have little impact on model performance. \\

To simulate a plume in a negative image, we use a physics based procedure based on the work of \cite{gorrono2023understanding}. This method simulates the per pixel transmittance of the plume in methane-absorbing bands of Sentinel-2 and Landsat (B11 and B12 for Sentinel-2 and B6 and B7 for Landsat). The input for this process is a real methane concentration image ($\Delta$CH$_4$) sampled from positive images in our dataset. This contrasts with other works that use synthetic plumes that do not produce realistic methane concentrations~\cite{zortea_detection_2023,rouet2024automatic}. Specifically, for an image without a plume (${\neg plume}$) we sample a plume from the training dataset and crop the methane concentration image ($\Delta$CH$_4$) with the plume mask. With the $\Delta$CH$_4$ cropped image and the viewing geometry and solar angles of the clear scene, we estimate the transmittance for bands 11 and 12 ($\tau_{B12}$, $\tau_{B11}$) (bands 6 and 7 of Landsat). The transmittance estimation is based on the MODTRAN radiative transfer model~\cite{MODTRAN}. We use MODTRAN to generate a look-up table (LUT) with the relationship between the methane enhancement ($\Delta$CH$_4$) and the transmittance at a fine spectral resolution $T_{\Delta\text{CH}_4}(\lambda)$ for different viewing geometries assuming a constant background concentration of 1800 ppb. For a given $\Delta$CH$_4$ and solar and viewing zenith angles, we use bicubic-spline interpolation of the LUT to obtain $T_{\Delta\text{CH}_4}(\lambda)$ which we then integrate over the spectral response function (SRF) of the satellite. The plume is then injected on the bands of the image following the equation~\cite{gorrono2023understanding}: 

\begin{align}
 BA &=  BA_{\neg plume} \frac{ \int E_g(\lambda) T_{atm}(\lambda) T_{\Delta \text{CH}_4}(\lambda) \text{srf}_{BA}(\lambda) d\lambda}{\int E_g(\lambda) T_{atm}(\lambda) \text{srf}_{BA}(\lambda) d\lambda}\\
   &= BA_{\neg plume} \cdot \tau_{BA}
\end{align}

\begin{figure}
    \centering
    \includegraphics[width=1.0\linewidth]{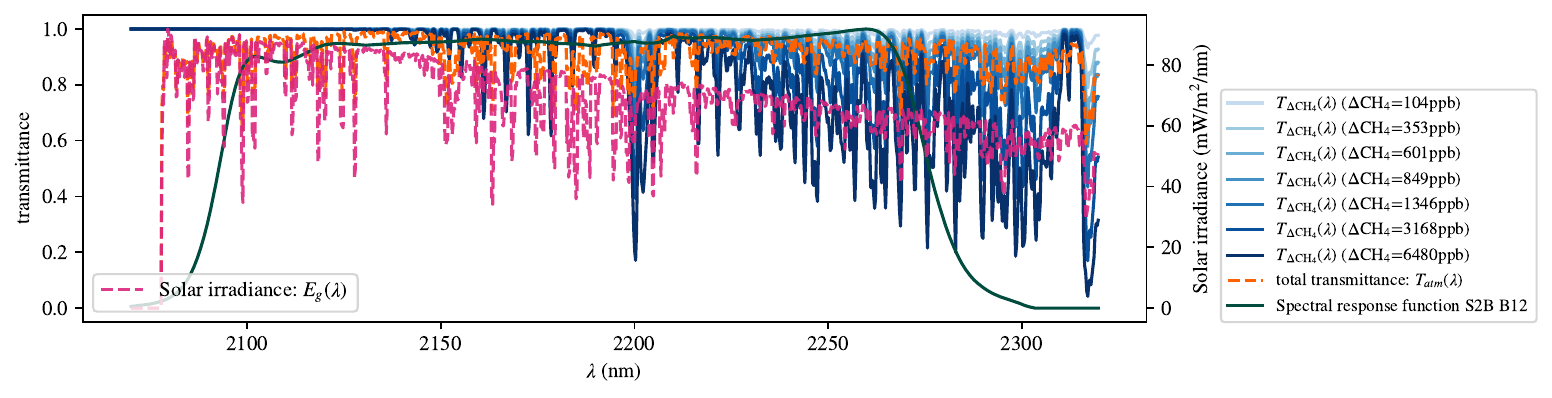}\\
\includegraphics[width=.6\linewidth]{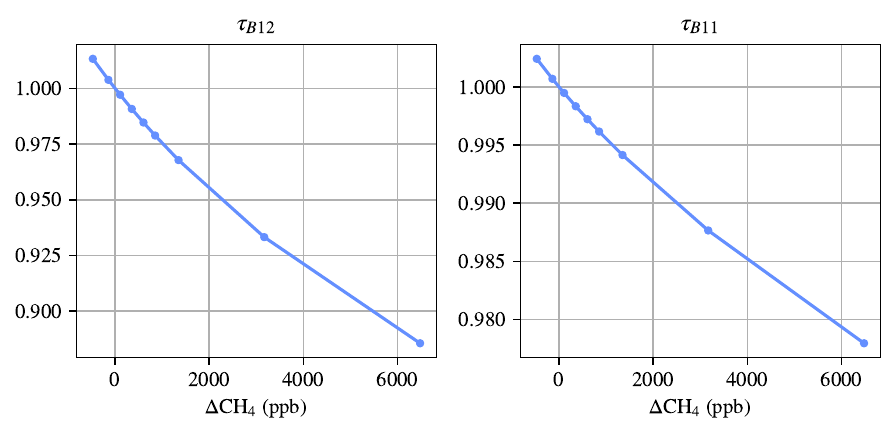}
    \caption{Top: Transmittance function for different  concentrations of methane ($\Delta$CH$_4$), spectral response function (SRF) of Sentinel-2B on the band 12, atmospheric transmittance $T_{atm}$ (gray) and solar irradiance $E_g$ (green). Bottom: Integrated transmittance for different  concentrations of methane for bands B11 and B12 of Sentinel-2. Figures obtained from MODTRAN radiative transfer simulations for an air mass factor of 3.22. }
    \label{fig:spectrum}
\end{figure}

Where $A$ refer to band 11 or 12 of Sentinel-2 or 6 or 7 of Landsat, $BA_{\neg plume}$ is the pixel value of the plume free image in the band, $T_{\Delta \text{CH}_4}(\lambda)$ is the transmittance for a concentration of $\Delta \text{CH}_4$ over the background, $E_g(\lambda)$ is the total solar irradiance, $T_{atm}(\lambda)$ is the atmospheric transmittance and the integral is taken over the SRF of the satellite for the band. We use an standard atmosphere simulation from LibRADTRAN for $E_g$ and $T_{atm}$ and assume constant surface reflectance over band wavelengths. Figure~\ref{fig:spectrum} shows on the top the different elements of this equation for different methane concentrations. On the bottom we display the integrated transmittances for different concentrations on bands B11 and B12 of Sentinel-2. \\

Wind plays a major role in detecting methane plumes. Higher winds disperse the plume more rapidly, making it difficult to detect. In addition wind direction provides evidence to distinguish weak plumes from artifacts. As weaker plumes are not visible at high wind speed, the simulation process takes into account the wind conditions of the clear image in order to inject the plume in the B11 and B12 bands. Specifically, for a clear image with wind speed $w$ we sample plumes from the training dataset with wind speed $w'$ such that $\|w-w'\|\leq 1.5$; however, if $w$ is higher than 9 m/s we do not simulate the plume. After sampling the plume, we rotate the methane concentration image to align it with the wind direction of the image to simulate into. We note that this \emph{wind consistent} procedure is required because the proposed detection model uses the wind field as input. \\

MARS-S2L is trained for 170 epochs of 65,536 samples each with Adam optimisation \cite{kingma2014adam}, a learning rate of 5e-4, weight regularisation of 1e-6 and early stopping. Model selection is performed using mean average precision on the validation split. Validation is always performed on the raw data with no simulation. 
 \\

\subsection*{Comparison to CH4Net}
For comparison to CH4Net, we retrain the model developed in \cite{vaughan2023ch4net} on the same training dataset utilised for MARS-S2L. As CH4Net was originally developed for Sentinel-2 only and utilised all 13 bands, we opt to retrain with only the five bands overlapping with Landsat to allow for application to this instrument and fair comparison with MARS-S2L on all images. CH4Net is trained for 200 epochs using the simulation scheme described above with Adam optimisation \cite{kingma2014adam}, a learning rate of 5e-4, weight regularisation of 1e-6 and early stopping.

\subsection*{Controlled release experiments}
As images are hand-labelled based on the MBMP images, we do not have a ground truth value for the total emission rate. To test the ability of MARS-S2L to detect small emissions, we run the model on all images from the Stanford controlled releases of 2021 and 2022~\cite{sherwin_single-blind_2023,sherwin_single-blind_2024}. In these experiments, methane was periodically released coinciding with Sentinel-2 and Landsat overpasses to test the detection capabilities of different satellites. \\

\begin{figure}
    \centering
    \includegraphics[width=0.63\linewidth]{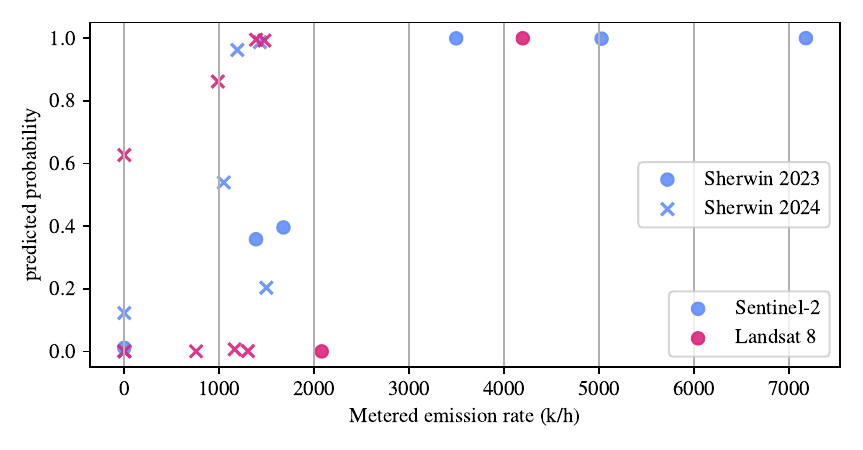}\includegraphics[width=0.33\linewidth]{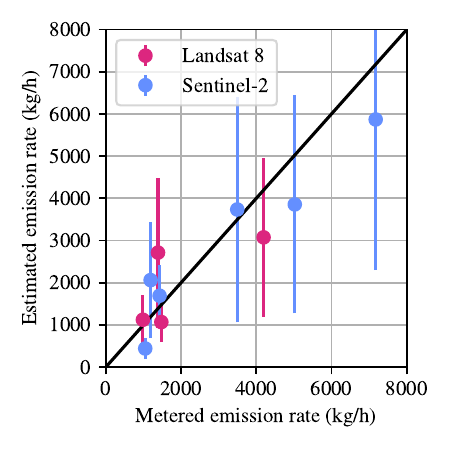}
    \caption{Left predicted probability of the model in the controlled release experiments of Sherwin et al.~\cite{sherwin_single-blind_2023,sherwin_single-blind_2024}. Right: quantified emissions of positive samples using the mask provided by the MARS-S2L model.}
    \label{fig:controlled_releases}
\end{figure}

Releases for both experiments were made at two different sites in Arizona: Ehrenberg \cite{sherwin_single-blind_2023} and near Casa Grande \cite{sherwin_single-blind_2024}. For the first set of releases \cite{sherwin_single-blind_2023}, we evaluated on the eight data points of the experiment (six from Sentinel-2 and two from Landsat), removing the Sentinel-2 acquisition from 24th October 2021 due to cloud cover. In total, the first experiment consisted of 7 positive and 1 negative images with releases up to 7.5 t/h. The second set of releases~\cite{sherwin_single-blind_2024},  were significantly more challenging, with 10 positive and 5 negative samples and release rates between 0.75 to 1.5 t/h. Figure~\ref{fig:controlled_releases} left shows the model output probability over all images in the experiment. We see that the model is able to detect all large emissions ($\ge$ 3 t/h) and produced only one false positive with medium probability (0.6). For small emissions (between 0.75 and 2 t/h) 4 are false negatives, 9 true positives and 4 are given medium probabilities (between 0.2 and 0.8). Results by satellite show higher disagreements in Landsat: 3 out of 4 false positives and 1 false negative. This is hypothesized to be due to the lower number of Landsat images in the training dataset (see Table~\ref{tab:splits}) and lower spatial resolution of this sensor.\\

On the right of Figure~~\ref{fig:controlled_releases} we show the estimated vs real emissions for the positive detections. For the estimation of the flux rate we use integrated mass enhancement method~\cite{gorrono2023understanding,varon2021high} with the plume mask produced by the model and the concentration image obtained from the quantification of the MBMP ratio~\cite{gorrono2023understanding}. Overall we see a good agreement similar to those obtained with manual delineations in the controlled release articles~\cite{sherwin_single-blind_2023,sherwin_single-blind_2024}.



\subsection*{Thailand platform timeseries}
\label{sec:mys_1}
\begin{figure}
    \centering
    \includegraphics[width=1.0\linewidth]{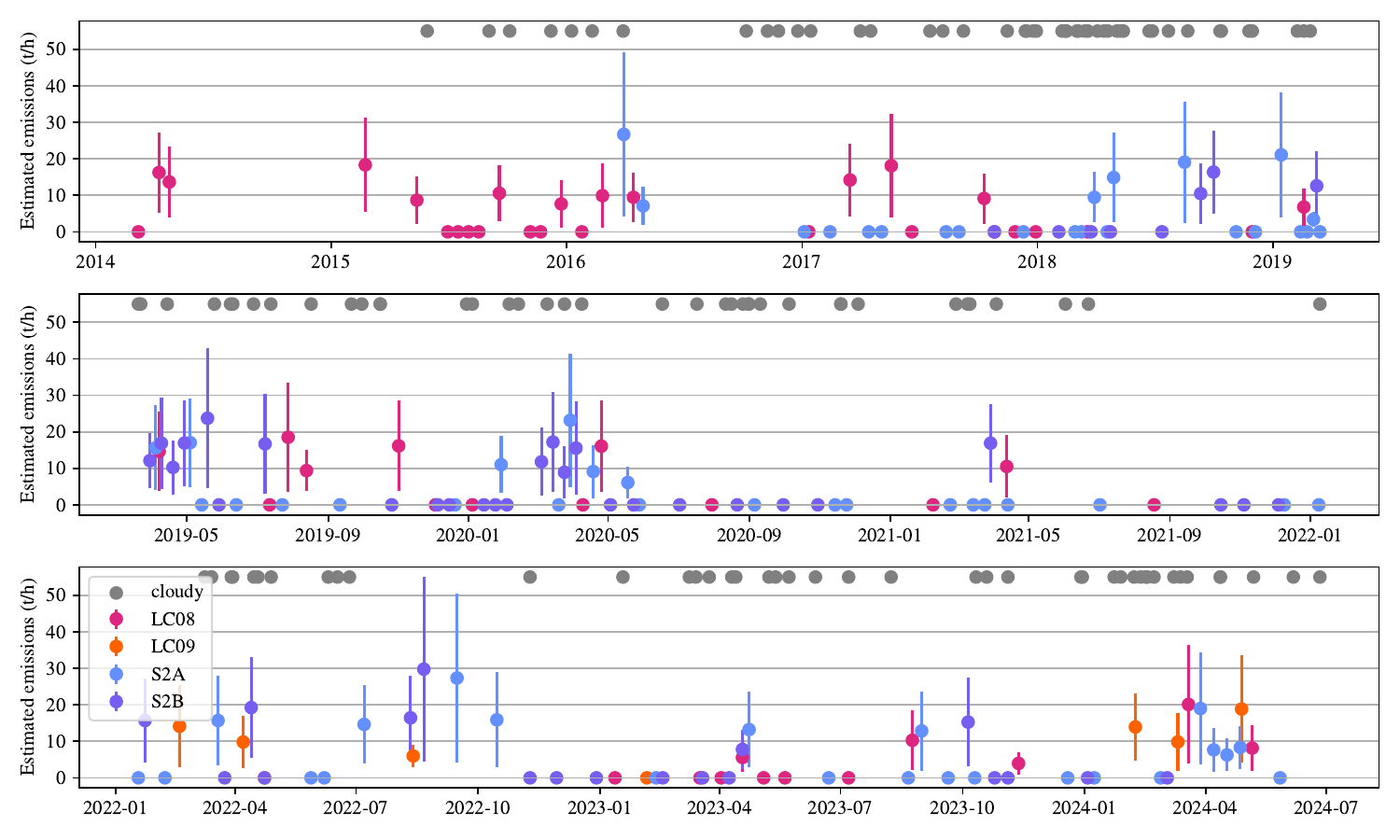}
    \caption{Full time series of validated and quantified emissions from the offshore platform in the Gulf of Thailand.}
    \label{fig:malaysia}
\end{figure}

\begin{figure}
    \centering
    \includegraphics[height=.7\paperheight]{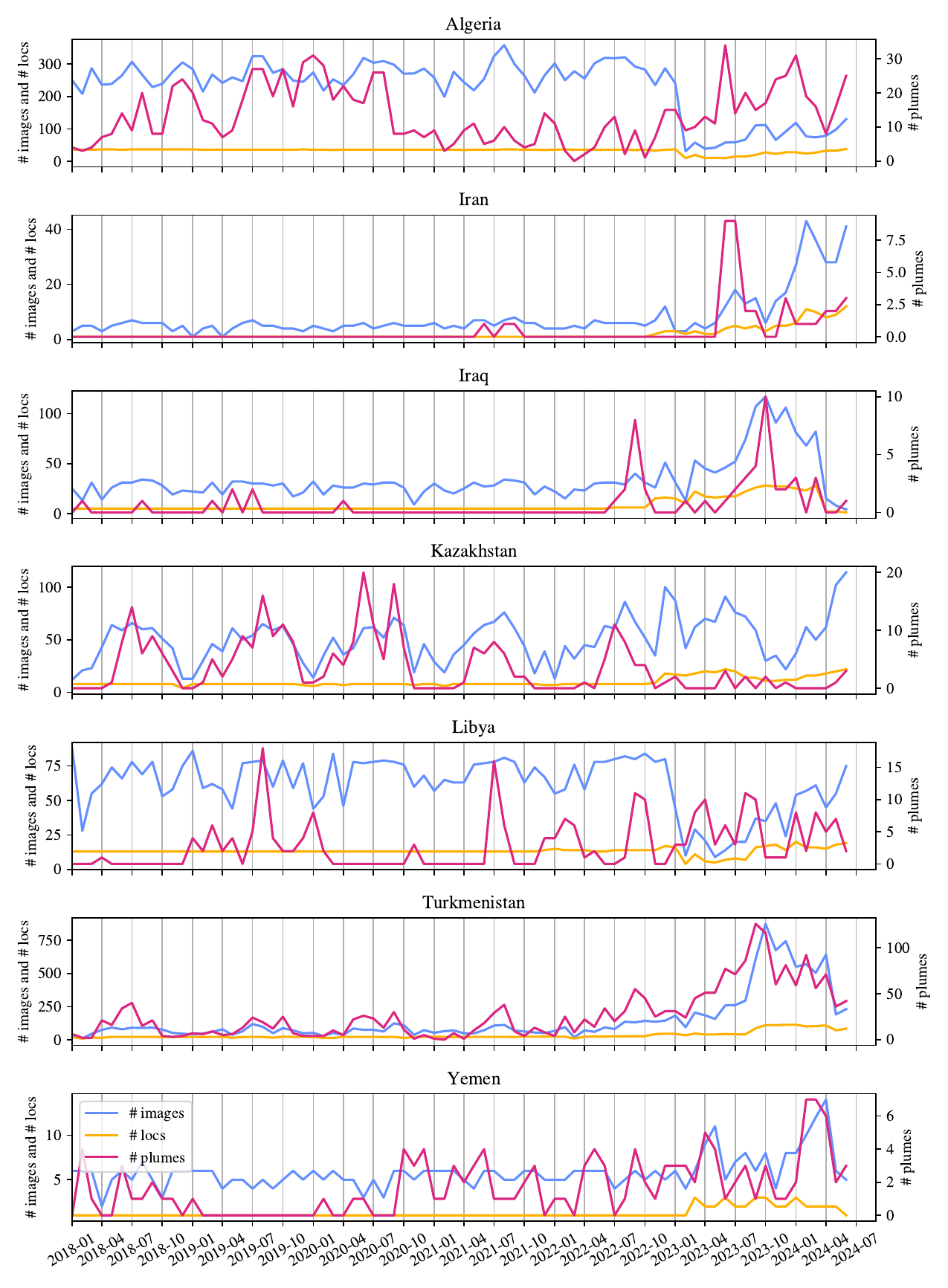}\\
    \caption{Monthly number of images and plumes by country.}
    \label{fig:nimagesplumesbycountrymars2}
\end{figure}

\begin{figure}
    \centering
    \includegraphics[height=.7\paperheight]{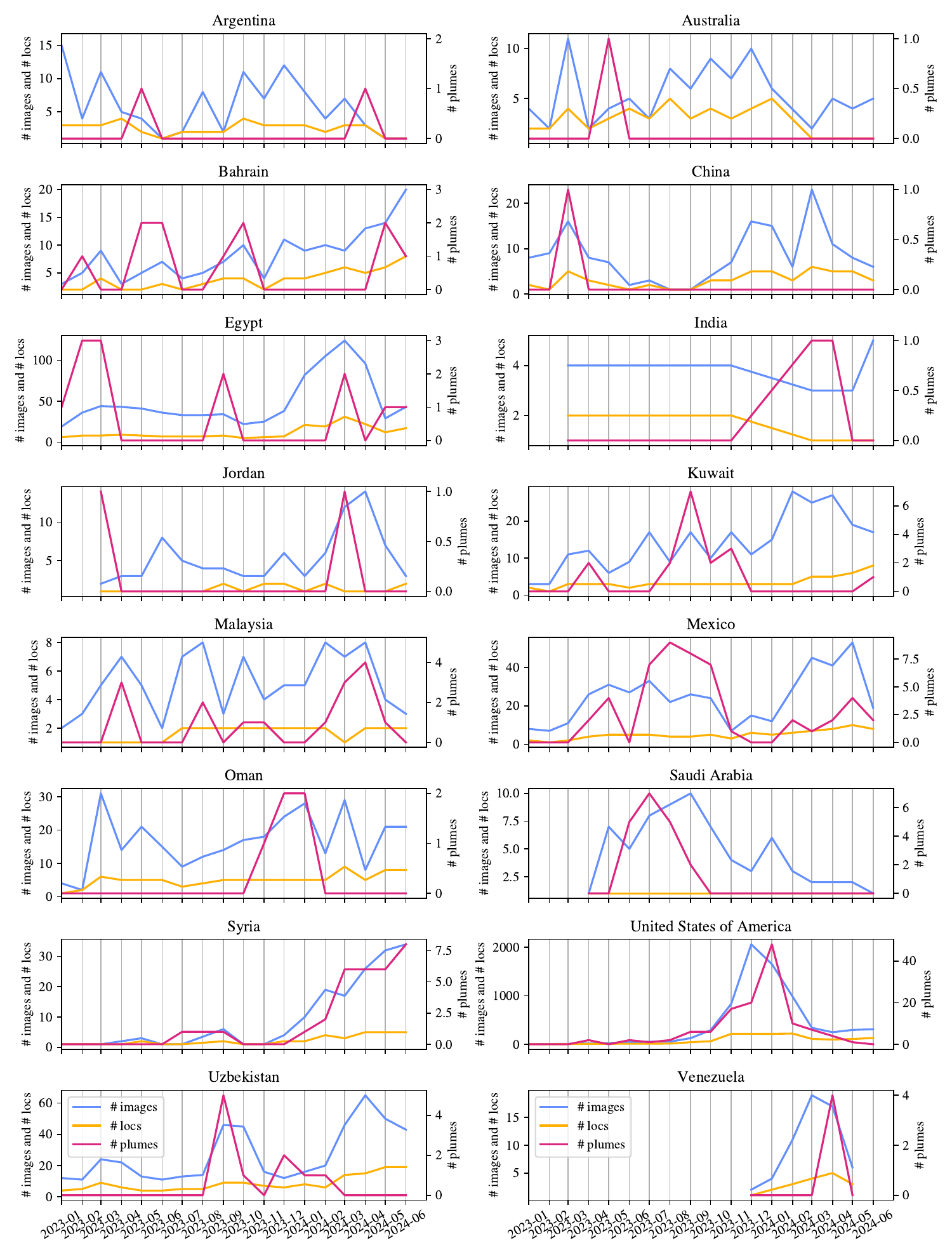}\\
    \caption{Monthly number of images and plumes by country.}
    \label{fig:nimagesplumesbycountryrest}
\end{figure}

A longitudinal timeseries of observed emissions from the offshore platform in the Gulf of Thailand discussed in the introduction is shown in Figure \ref{fig:malaysia}. This work was generated as part of an IMEO case study of this site in 2024 \cite{clark2024big}. Out of the 310 Landsat and Sentinel-2 observations, 129 are marked as invalid or cloudy, 109 as no plume observed and 72 as positive plumes. All these detections have been manually verified by UNEP IMEO analysts. 

\begin{table}
\centering
\begin{tabular}{lrrr}
\toprule
 & images & plumes & sites \\
country &  &  &  \\
\midrule
Algeria & 17824 & 1064 & 50 \\
Angola & 12 & 0 & 2 \\
Argentina & 161 & 6 & 5 \\
Australia & 101 & 1 & 7 \\
Azerbaijan & 85 & 0 & 5 \\
Bahrain & 231 & 21 & 8 \\
Canada & 1 & 0 & 1 \\
China & 160 & 1 & 7 \\
Egypt & 895 & 14 & 43 \\
Germany & 2 & 0 & 1 \\
India & 22 & 2 & 3 \\
Iran & 628 & 38 & 15 \\
Iraq & 2645 & 54 & 32 \\
Italy & 4 & 0 & 1 \\
Jordan & 86 & 2 & 2 \\
Kazakhstan & 3958 & 283 & 32 \\
Kuwait & 263 & 18 & 8 \\
Libya & 4810 & 227 & 27 \\
Malaysia & 332 & 72 & 2 \\
Mexico & 642 & 60 & 12 \\
Mozambique & 31 & 0 & 1 \\
Nigeria & 13 & 0 & 1 \\
Oman & 308 & 5 & 10 \\
Pakistan & 56 & 0 & 4 \\
Poland & 14 & 2 & 2 \\
Russia & 6 & 1 & 1 \\
Saudi Arabia & 68 & 19 & 1 \\
Syria & 169 & 36 & 6 \\
Turkmenistan & 11753 & 2198 & 122 \\
United States of America & 7361 & 126 & 262 \\
Uzbekistan & 562 & 13 & 25 \\
Venezuela & 59 & 4 & 5 \\
Yemen & 453 & 130 & 3 \\
\bottomrule
\end{tabular}

\caption{Test dataset statistics by country, showing from left to right total number of plumes, total images, and total sites.}
\label{tab:stats}
\end{table}

\begin{table}
\centering

\begin{tabular}{lrrr}
\toprule
 & images & plumes & sites \\
country &  &  &  \\
\midrule
Algeria & 669 & 140 & 45 \\
Angola & 12 & 0 & 2 \\
Argentina & 36 & 1 & 3 \\
Australia & 43 & 0 & 7 \\
Azerbaijan & 55 & 0 & 5 \\
Bahrain & 86 & 3 & 8 \\
Canada & 1 & 0 & 1 \\
China & 85 & 0 & 7 \\
Egypt & 517 & 4 & 43 \\
Germany & 2 & 0 & 1 \\
India & 14 & 2 & 1 \\
Iran & 220 & 13 & 15 \\
Iraq & 364 & 9 & 30 \\
Italy & 4 & 0 & 1 \\
Jordan & 51 & 1 & 2 \\
Kazakhstan & 449 & 5 & 24 \\
Kuwait & 142 & 1 & 8 \\
Libya & 371 & 33 & 26 \\
Malaysia & 40 & 9 & 2 \\
Mexico & 214 & 11 & 12 \\
Nigeria & 12 & 0 & 1 \\
Oman & 144 & 4 & 10 \\
Pakistan & 56 & 0 & 4 \\
Poland & 1 & 0 & 1 \\
Russia & 6 & 1 & 1 \\
Saudi Arabia & 17 & 0 & 1 \\
Syria & 142 & 29 & 5 \\
Turkmenistan & 3430 & 437 & 122 \\
United States of America & 5891 & 90 & 259 \\
Uzbekistan & 252 & 4 & 25 \\
Venezuela & 59 & 4 & 5 \\
Yemen & 63 & 28 & 3 \\
\bottomrule
\end{tabular}
\caption{Test dataset statistics by country, showing from left to right total number of plumes, total images, and total sites.}
\label{tab:stats_test}
\end{table}

\begin{table}[]
    \centering
    \begin{tabular}{llrrrll}
\toprule
 &  & total  & total & total & min date & max date \\
split & Satellite & plumes & images & sites &  &  \\
\midrule
\multirow[t]{2}{*}{Train} & Landsat & 757 & 4667 & 447 & 2018-01-05 10:02 & 2023-11-30 07:16 \\
 & Sentinel-2 & 2366 & 29414 & 525 & 2018-01-01 09:13 & 2023-11-30 07:12 \\
\cline{1-7}
\multirow[t]{2}{*}{Val} & Landsat-8/9 & 20 & 120 & 22 & 2021-01-07 07:22 & 2021-12-29 06:52 \\
 & Sentinel-2 & 258 & 5660 & 86 & 2021-01-01 07:03 & 2021-12-31 10:03 \\
\cline{1-7}
\multirow[t]{2}{*}{Test} & Landsat & 332 & 5515 & 663 & 2023-12-01 07:07 & 2024-06-29 18:16 \\
 & Sentinel-2 & 497 & 7933 & 648 & 2023-12-01 06:42 & 2024-06-29 10:20 \\
\bottomrule
\end{tabular}
    \caption{Training, test and validation dataset statistics.}
    \label{tab:splits}
\end{table}

\section*{PlumeViewer}
\label{sec:plumeviewer}
\begin{figure}
    \centering
    \includegraphics[height=.35\paperheight]{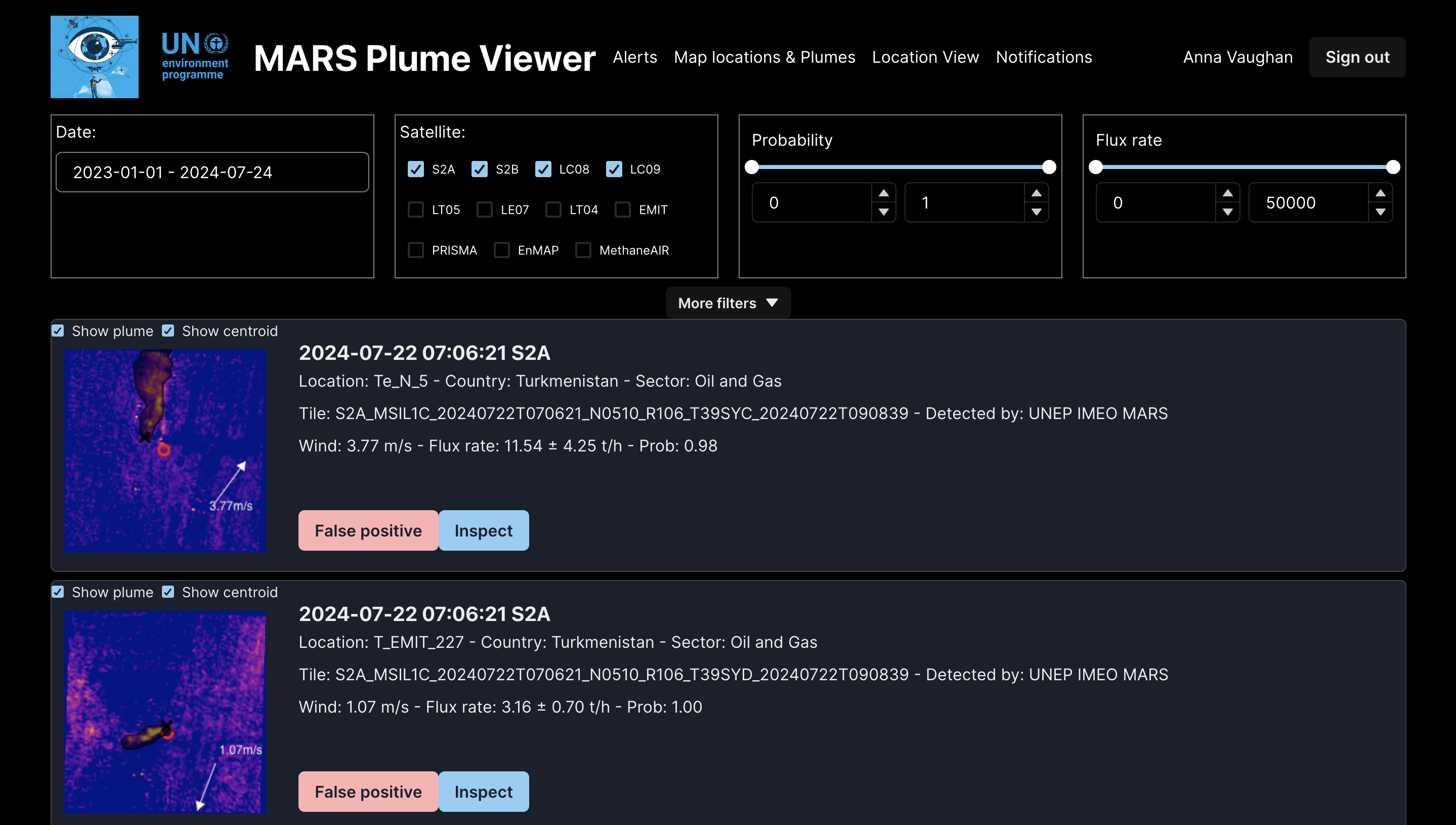}\\
    \caption{PlumeViewer alert view showing model predictions ordered from highest to lowest predicted probability. }
    \label{fig:plumeviewer1}
\end{figure}

\begin{figure}
    \centering
    \includegraphics[height=.35\paperheight]{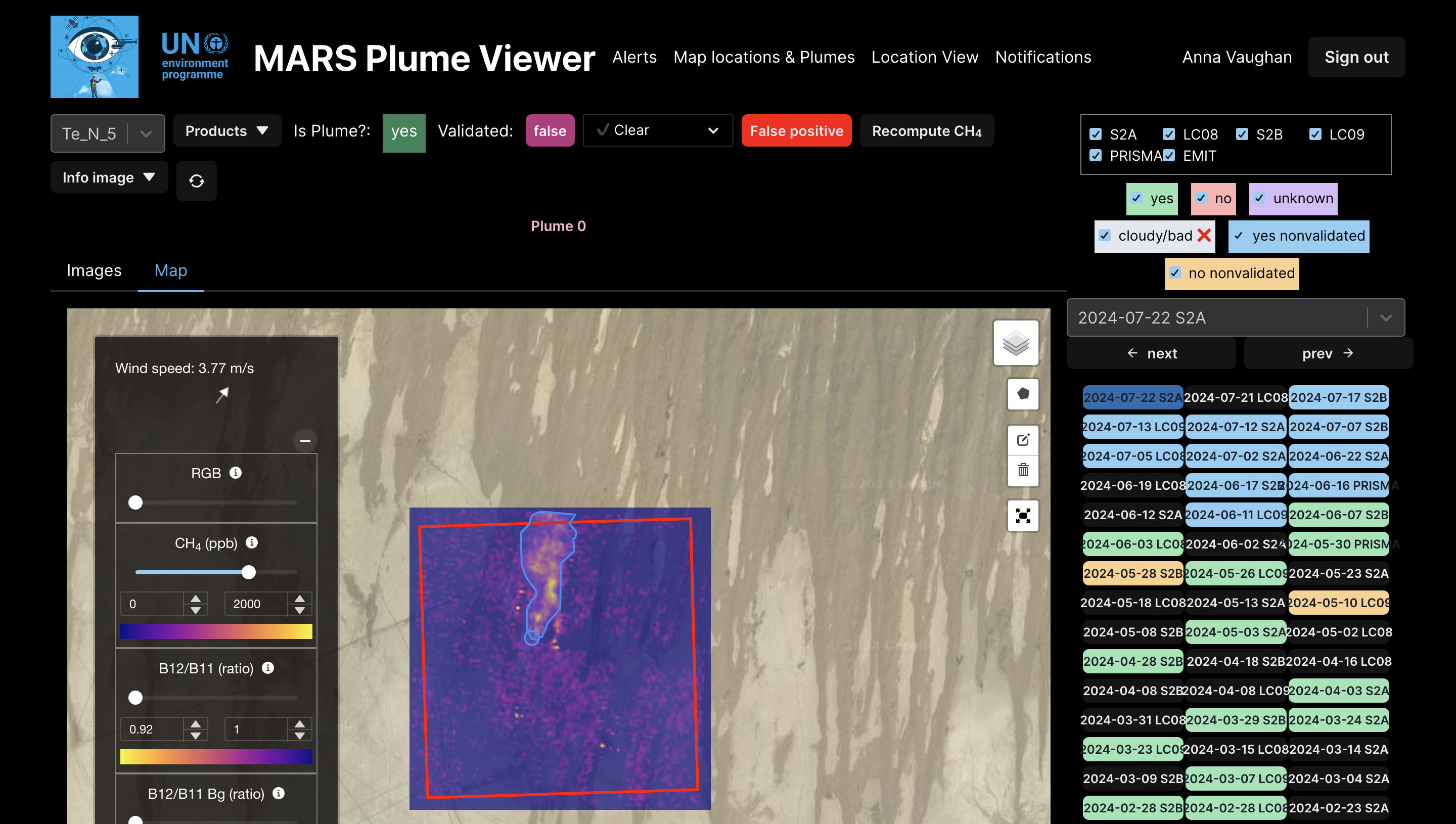}\\
    \caption{PlumeViewer location view showing a model alert for a site.}
    \label{fig:plumeviewer2}
\end{figure}

\begin{figure}
    \centering
    \includegraphics[height=.35\paperheight]{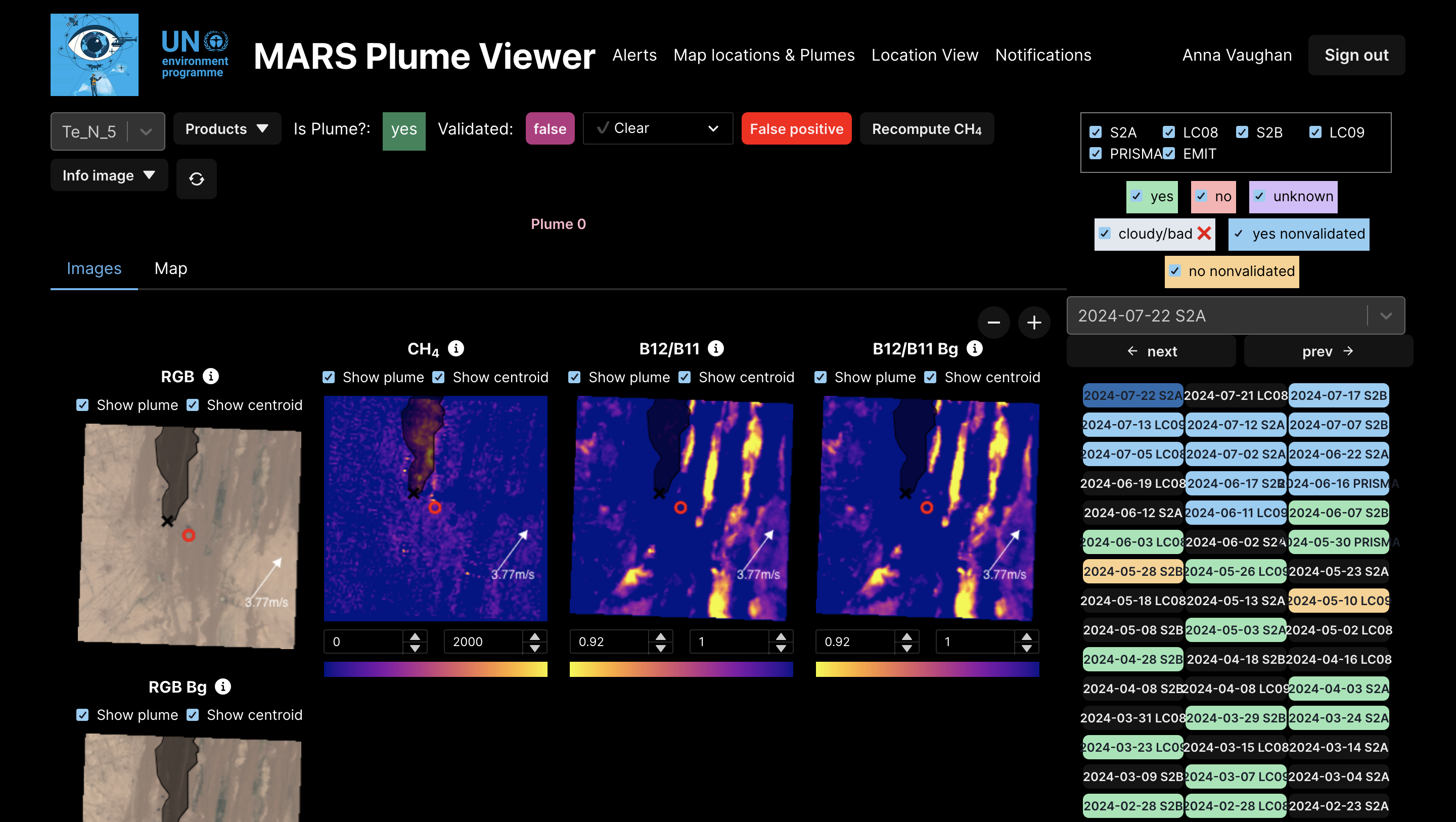}\\
    \caption{PlumeViewer location view showing a auxiliary images for a site.}
    \label{fig:plumeviewer3}
\end{figure}

Figures \ref{fig:plumeviewer1}, \ref{fig:plumeviewer2} and \ref{fig:plumeviewer3} show the three stages of an analyst inspecting an alert produced by MARS-S2L. Figure \ref{fig:plumeviewer1} shows the alert screen where the analyst can inspect model predictions. Any of these alerts can be selected for verification (Figure \ref{fig:plumeviewer2}) with multiple different auxiliary images available for analysis (Figure \ref{fig:plumeviewer3}). 

\subsection*{Model prediction examples}
Example timeseries of model predictions are shown in Figures S11-S18 for emitters in Argentina, Thailand, Bahrain, Libya, the US, Algeria, Turkmenistan and Yemen. These figures showcase model performance across a range of background and emitter types in a diverse range of regions globally. 

\begin{figure}
    \centering
    \includegraphics[height=.35\paperheight]{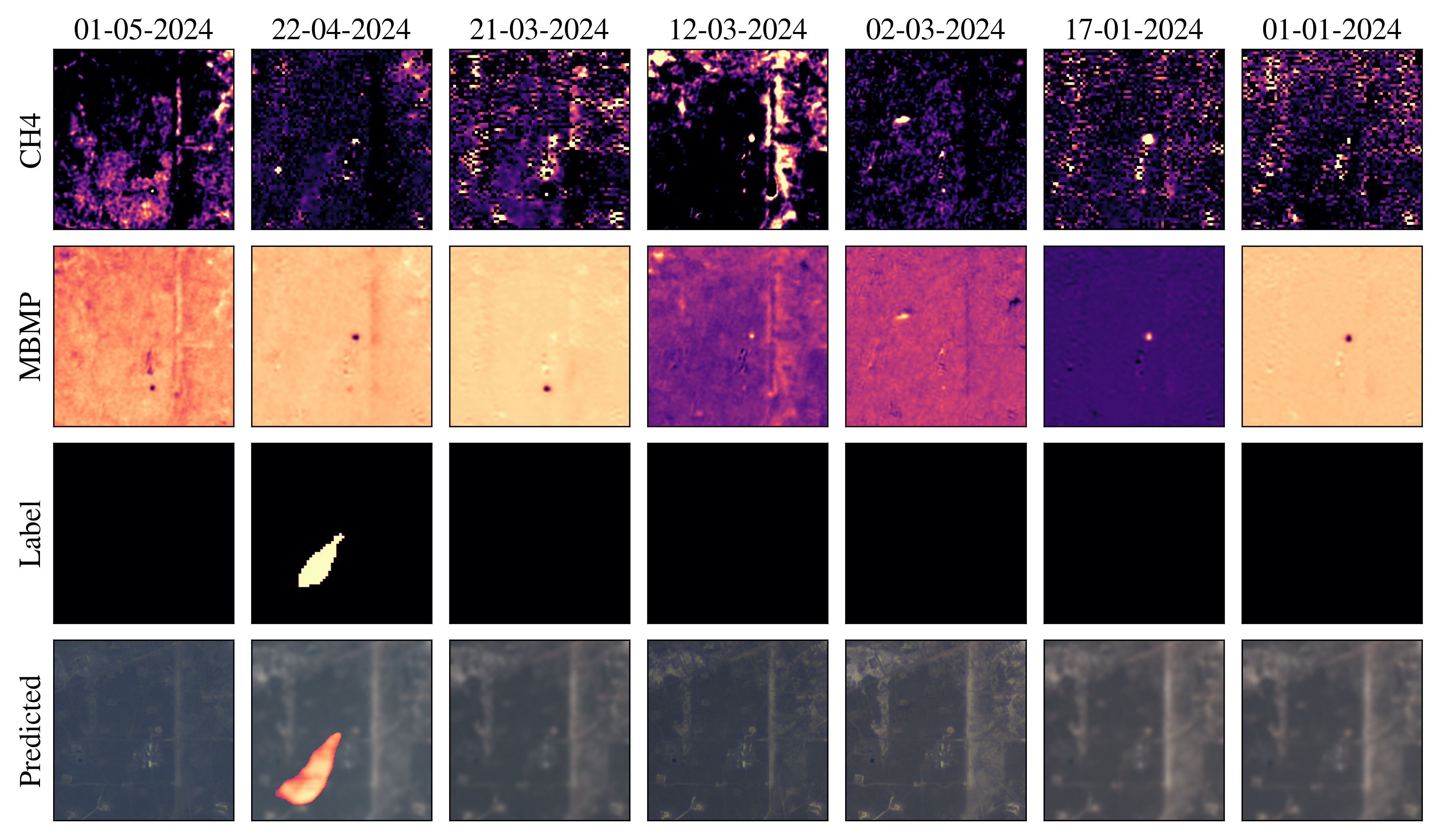}\\
    \caption{MARS-S2L predictions for an emitter in Argentina.}
    \label{fig:}
\end{figure}

\begin{figure}
    \centering
    \includegraphics[height=.35\paperheight]{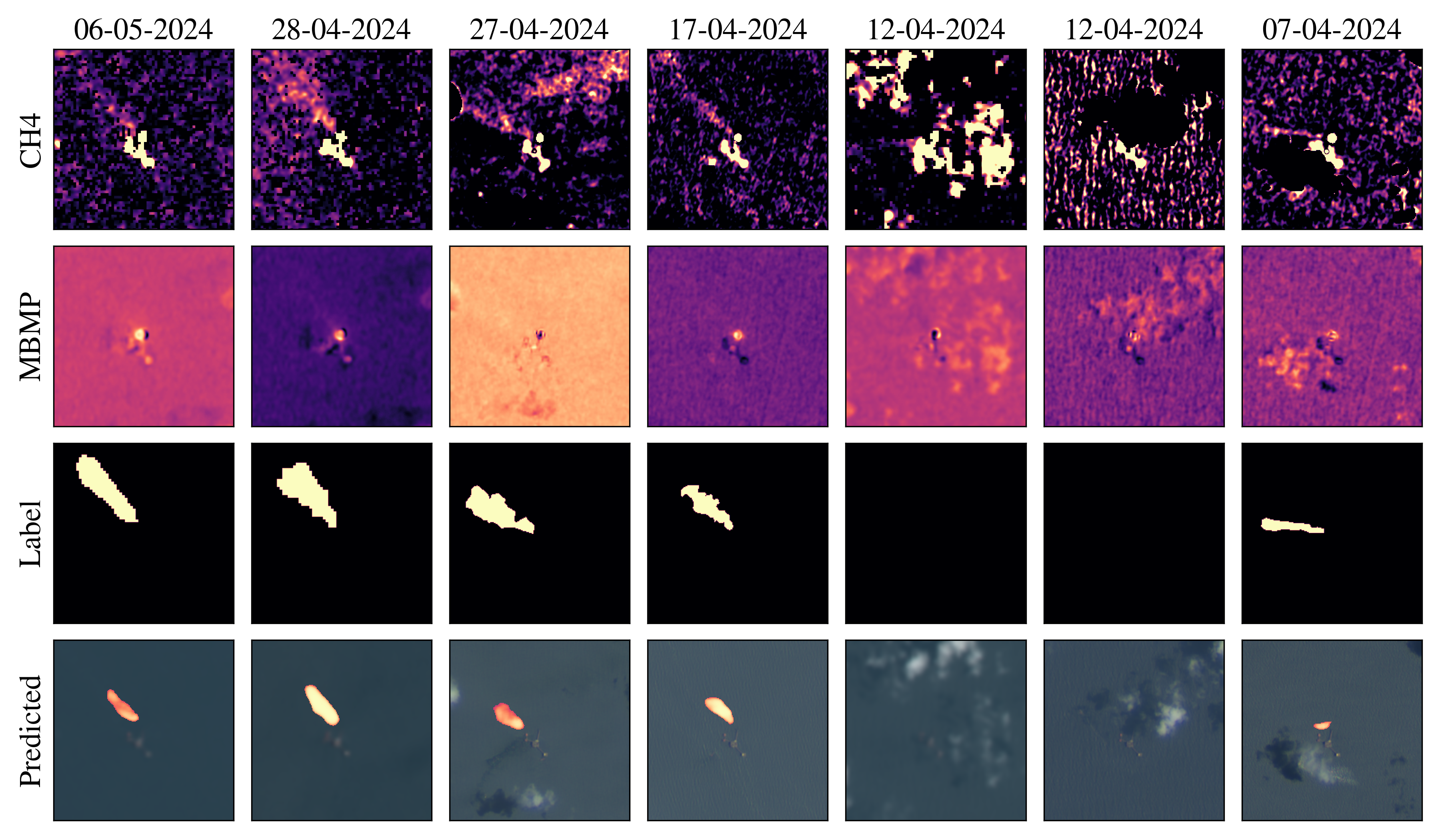}\\
    \caption{MARS-S2L predictions for an emitter in Thailand.}
    \label{fig:}
\end{figure}

\begin{figure}
    \centering
    \includegraphics[height=.35\paperheight]{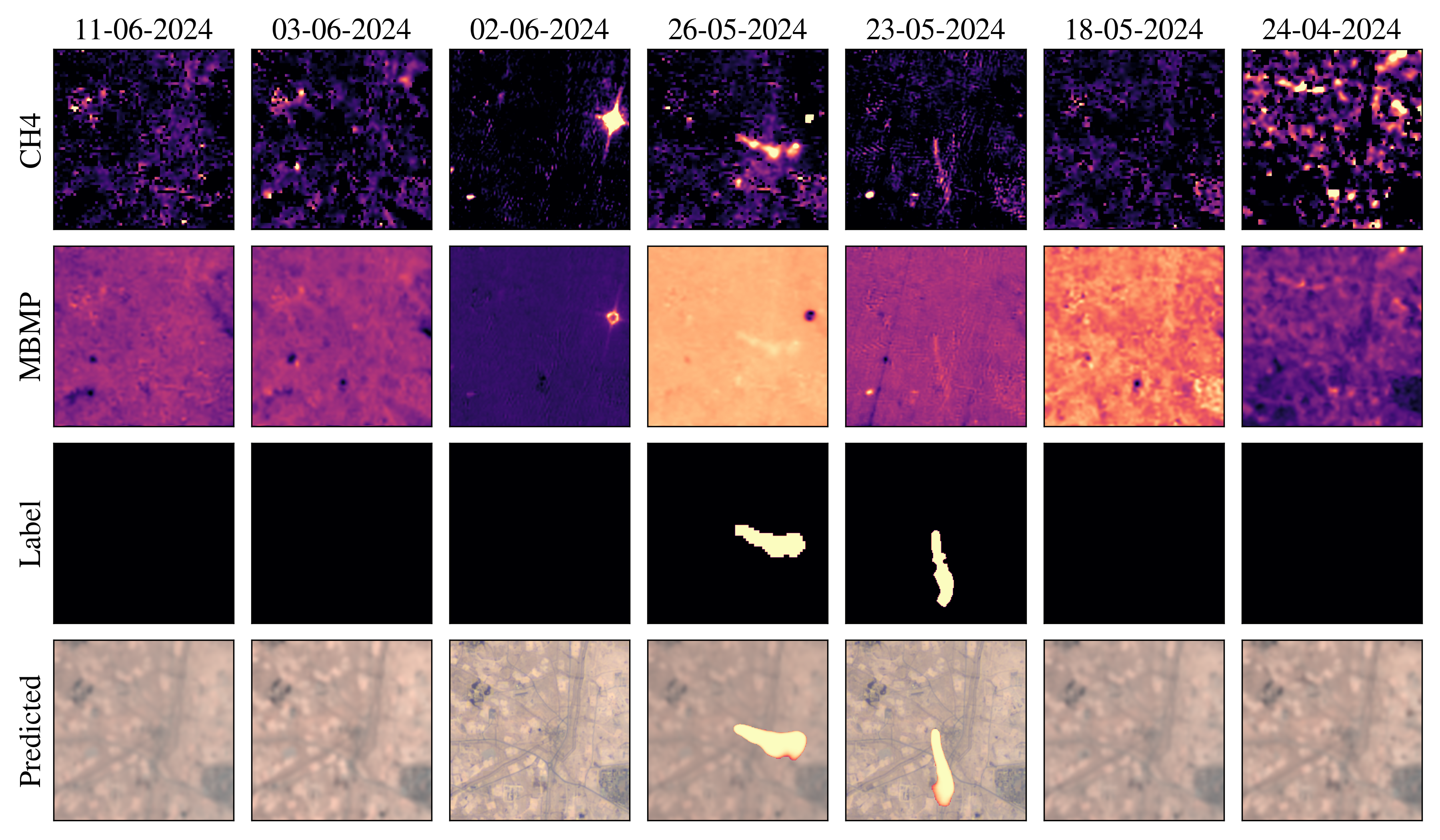}\\
    \caption{MARS-S2L predictions for an emitter in Bahrain.}
    \label{fig:}
\end{figure}

\begin{figure}
    \centering
    \includegraphics[height=.35\paperheight]{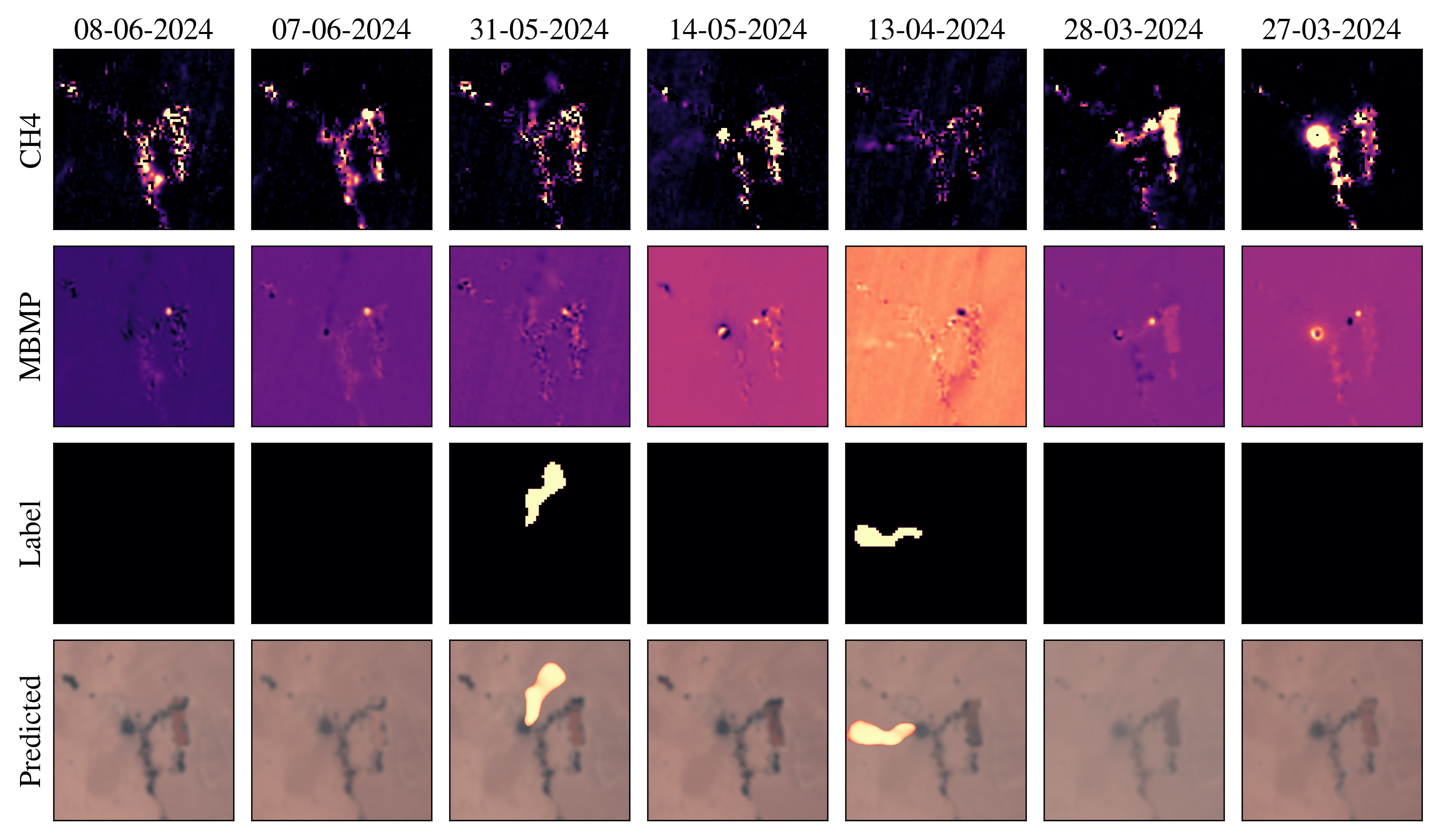}\\
    \caption{MARS-S2L predictions for an emitter in Libya.}
    \label{fig:}
\end{figure}

\begin{figure}
    \centering
    \includegraphics[height=.35\paperheight]{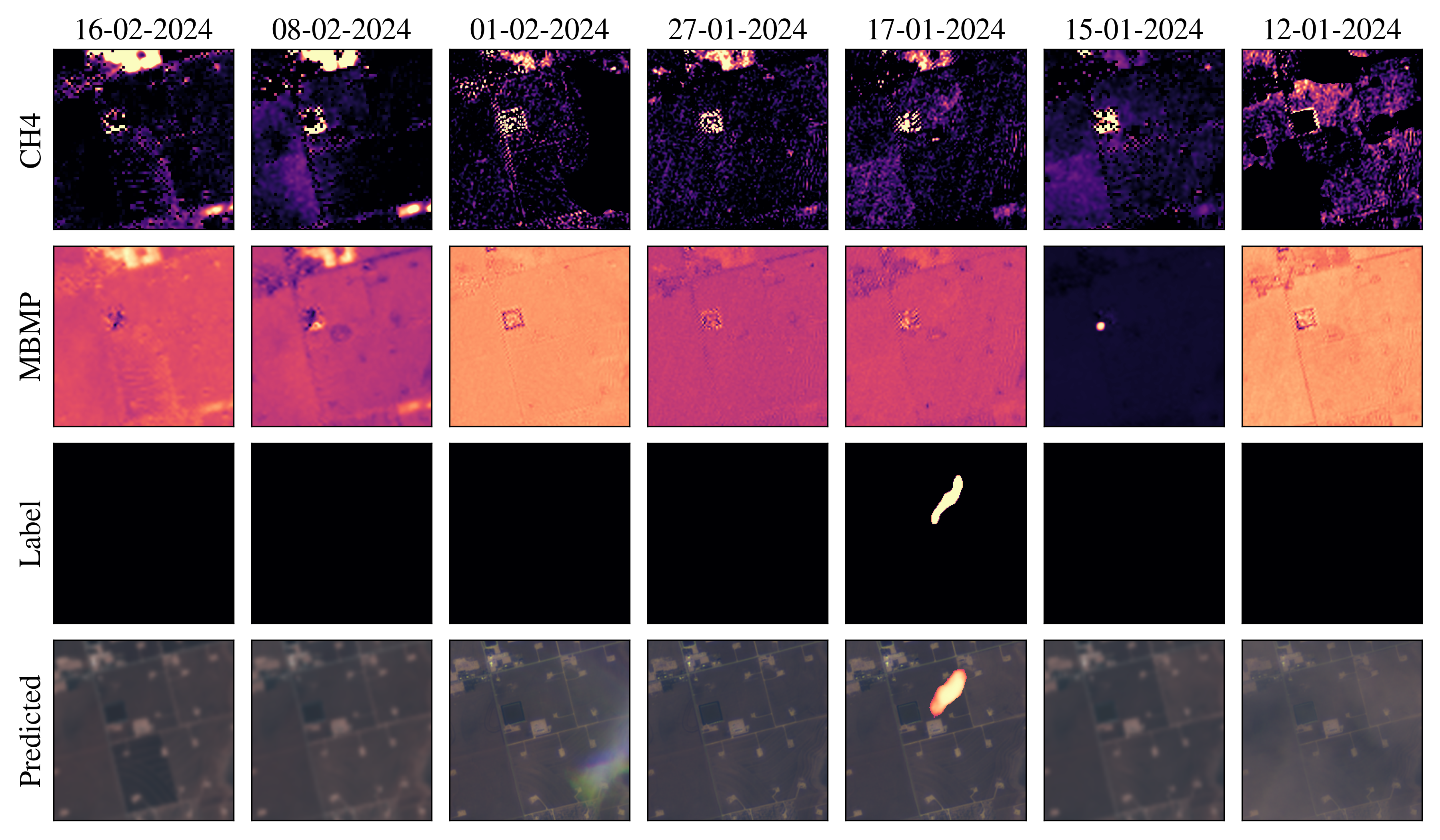}\\
    \caption{MARS-S2L predictions for an emitter in the US.}
    \label{fig:}
\end{figure}

\begin{figure}
    \centering
    \includegraphics[height=.35\paperheight]{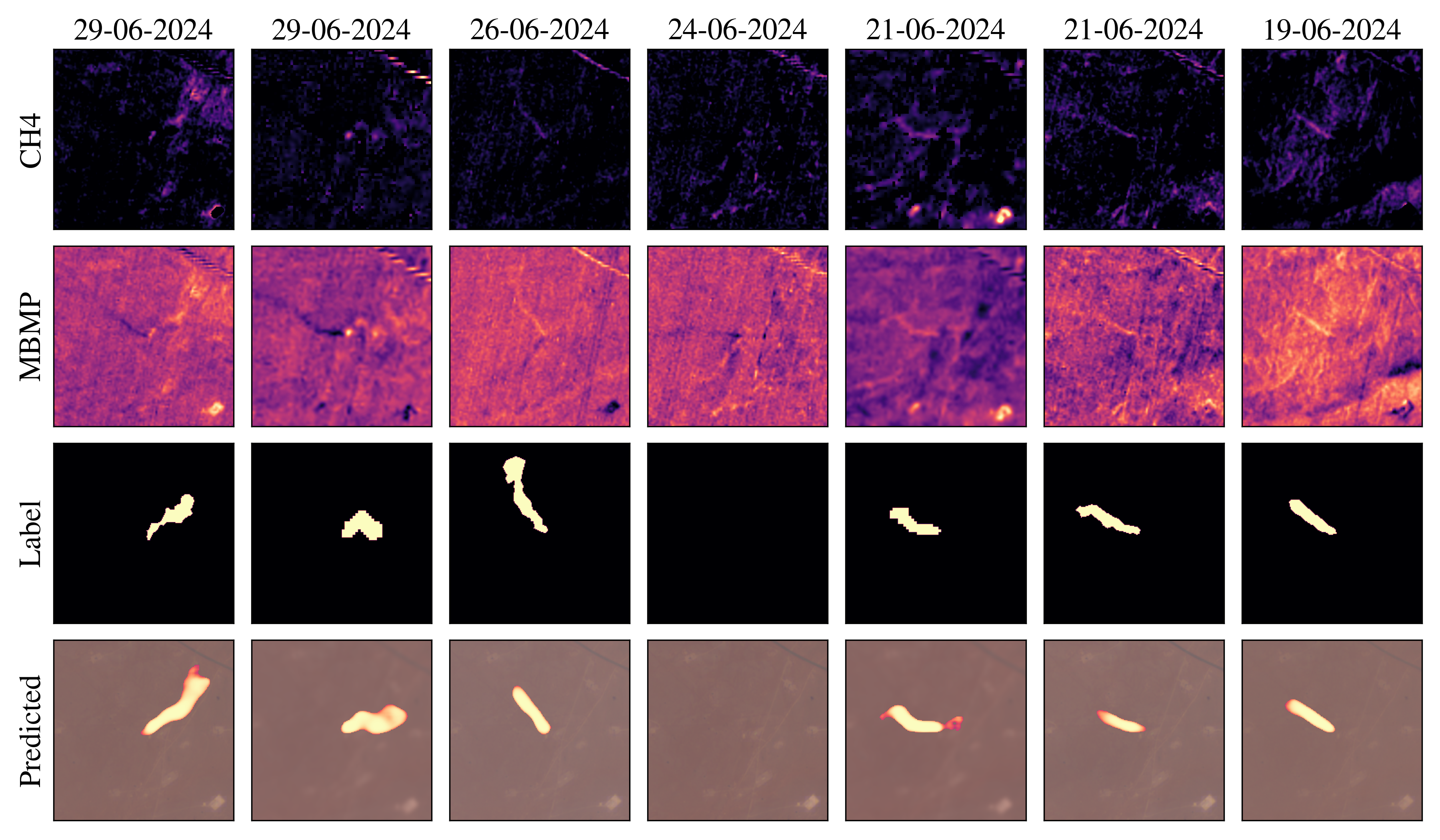}\\
    \caption{MARS-S2L predictions for an emitter in Algeria.}
    \label{fig:}
\end{figure}

\begin{figure}
    \centering
    \includegraphics[height=.35\paperheight]{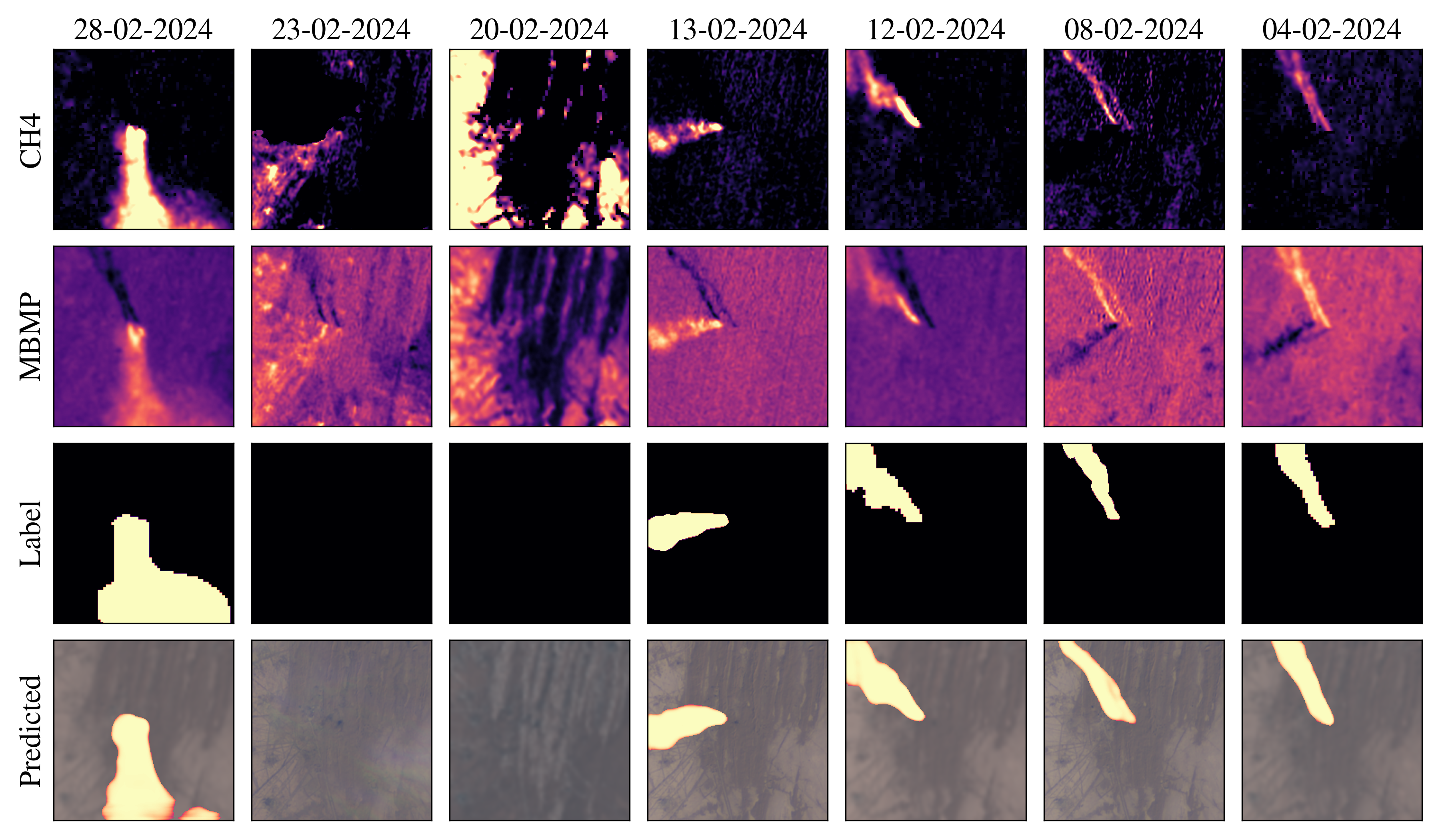}\\
    \caption{MARS-S2L predictions for an emitter in Turkmenistan.}
    \label{fig:}
\end{figure}

\begin{figure}
    \centering
    \includegraphics[height=.35\paperheight]{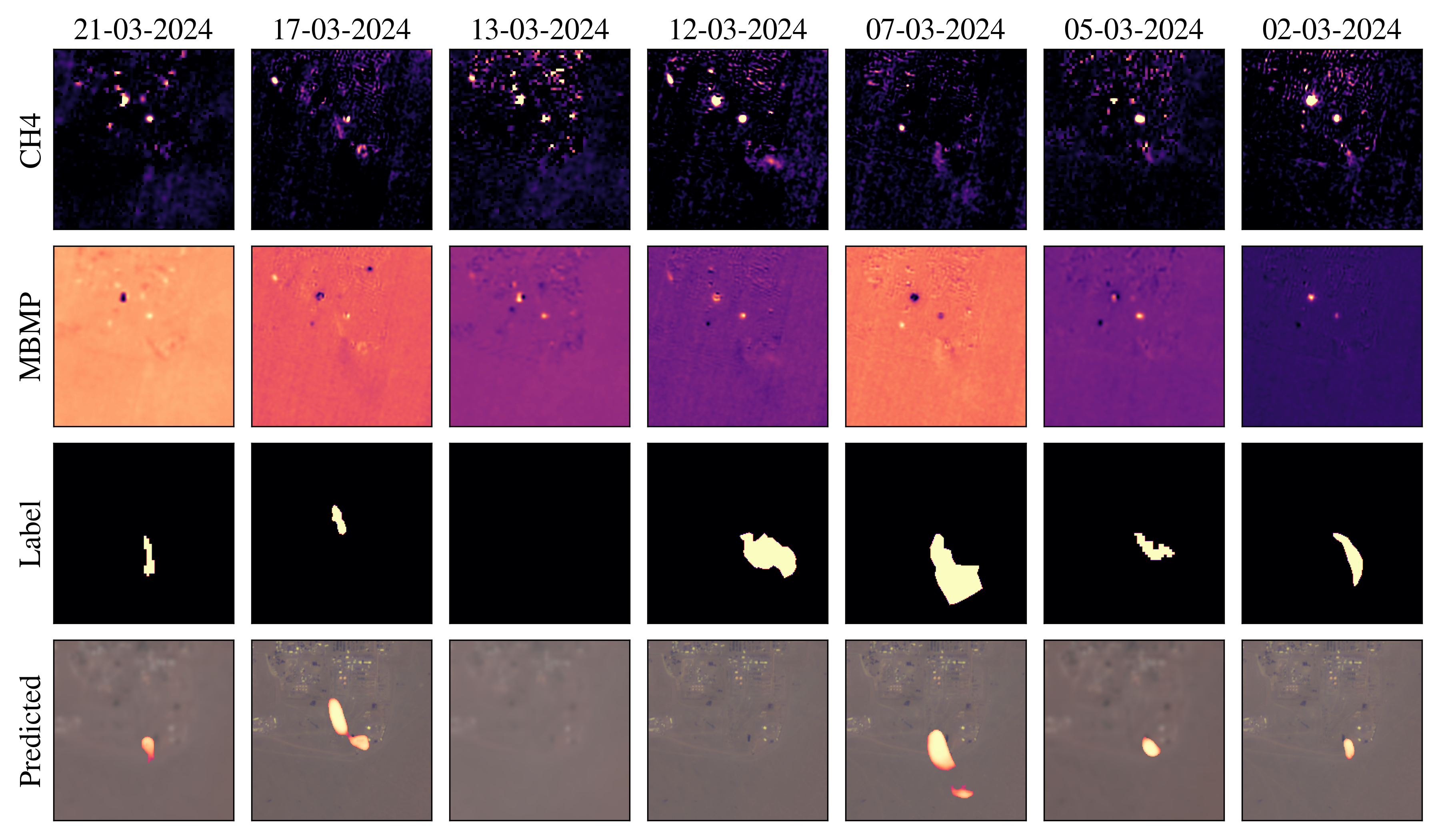}\\
    \caption{MARS-S2L predictions for an emitter in Yemen.}
    \label{fig:}
\end{figure}

\end{document}